\documentclass[a4paper,fleqn]{cas-dc}

\usepackage[numbers]{natbib}
\usepackage{algorithm}
\usepackage{algpseudocode}
\usepackage{subcaption}
\usepackage{float}
\usepackage{placeins} 

\def\tsc#1{\csdef{#1}{\textsc{\lowercase{#1}}\xspace}}
\tsc{WGM}
\tsc{QE}
\tsc{EP}
\tsc{PMS}
\tsc{BEC}
\tsc{DE}

\begin{document}
\let\WriteBookmarks\relax
\def\floatpagepagefraction{1}
\def\textpagefraction{.001}
\shorttitle{Uncertainty-aware AI-enabled framework for predicting hydrogen dispersion}
\shortauthors{Niclas Flehmig et~al.}

\title [mode = title]{Uncertainty-aware AI-enabled framework for predicting hydrogen dispersion in confined spaces to enhance operator situational awareness}                      

\author[1]{Niclas Flehmig}[orcid=0009-0006-6458-2689]
\cormark[1]
\ead{niclas.flehmig@ntnu.no}

\credit{Writing -- review \& editing, Writing -- original draft, Methodology, Investigation, Formal analysis, Data curation, Validation, Visualization, Conceptualization}

\affiliation[1]{organization={Department of Mechanical and Industrial Engineering, Norwegian University of Science and Technology},
                city={Trondheim},
                country={Norway}}

\author[2]{Mary Ann Lundteigen}

\credit{Writing -- review \& editing, Supervision, Methodology, Conceptualization}

\author[1]{Shen Yin}

\credit{Writing -- review \& editing, Supervision}

\affiliation[2]{organization={Department of Cybernetics Engineering},
                city={Trondheim},
                country={Norway}}

\cortext[cor1]{Corresponding author}

\begin{abstract}
We present a two-stage AI-enabled framework for enhancing operator situational awareness during compressed hydrogen bunkering in maritime confined spaces. In the event of a leak, the framework first estimates the unknown leakage rate from early sensor readings and then forecasts how the hydrogen cloud will disperse over the next time steps, providing operators with a predicted concentration map together with confidence bounds. The framework is based on two independent AI models building one AI system. The AI system is designed as a decision-support tool. It augments sensor readings, and thus enhances operator situational awareness and supports decision-making. This enables preventive safety barriers to be deployed before hazardous concentrations are reached, while an independent emergency shutdown system remains the ultimate safety layer. The system design incorporates safeguards to ensure the AI system operates within its competence envelope. Trained and tested on an open dataset of confined-space hydrogen releases, the framework achieves a mean absolute error of $0.052$~g/s in leakage-rate estimation and $\approx0.018$ in hydrogen-concentration prediction, approaching the sensor precision bound of $\pm0.01$. The hydrogen-concentration prediction provides well-calibrated uncertainty estimates that allow operators to judge the reliability of each forecast. These results suggest that the framework may provide early information about hazardous concentration build-up, potentially supporting preventive actions by operators.
\end{abstract}



\maketitle

\section{Introduction} \label{sec1}

\subsection{Motivation and Background} \label{sec11}

Green hydrogen is a potential candidate for decarbonizing maritime transport \cite{al-asmakhAlternativeFuelsDesign2025, yangReviewServicesManagement2026, ningRecentDevelopmentDecarbonization2023, zhakaHydrogenFuelMaritime2024}. Its gravimetric energy density is high and its combustion produces only water vapor. Yet the bunkering process, transferring fuel from a shore facility to a vessel's onboard tank, introduces significant safety challenges. This work focuses on shore-to-ship compressed $H_2$ bunkering, the recommended configuration for this fuel type \cite{handbookHydrogenVessels}.

The physical properties of hydrogen increase the risk of leaks. It has the smallest molecule, enabling permeation through seals and embrittlement of steel. It is flammable over a wide range (4--75\% in air) and ignites at very low energy. It is colorless, odorless, and buoyant, making leaks hard to detect and manage in enclosed spaces such as the tank connection space within a ship \cite{handbookHydrogenVessels, liuHydrogenRefuelingStations2025}. Currently, no dedicated maritime standard governs the system design of $H_2$ bunkering installations. Preventive and mitigation measures are therefore drawn from guidelines \cite{handbookHydrogenVessels}. These address design of high-pressure piping, open-deck station placement for natural ventilation, and dry break-away couplings.

The ultimate safety layer is the emergency shutdown (ESD), which triggers automatically when the lower flammability limit (LFL) is reached. An ESD is costly, operationally disruptive, and represents the last barrier before a potential catastrophic event. An early-warning system that forecasts hazardous concentration build-up before the LFL is reached would allow operators to trigger preventive safety barriers and thereby avoid the ESD.

Existing hydrogen sensors display only current concentration. They offer no predictive capability. Computational fluid dynamics (CFD) can model dispersion with high physical fidelity, but a single simulation requires hours of computation and expert configuration to set up the geometry, mesh, and boundary conditions. In a bunkering operation, where a leak evolves over seconds and operator decisions must be made within minutes, CFD is inherently incompatible with real-time response. AI systems can learn the underlying dispersion patterns from observational data and then generate forecasts within seconds, though at the cost of precision. This computational speed makes them suitable for real-time operation. The model can estimate the leakage rate and compute the predicted concentration field. This gives operators a forward-looking view of the hazard rather than a retrospective snapshot. However, their integration into safety-critical bunkering systems must be cautious. The AI system should augment sensor readings, and thus enhance operator situational awareness and support preventive action without undermining the intrinsic safety guaranteed by the independent emergency shutdown system.

\subsection{Related Work} \label{sec12}
The preceding review is based on a brief, targeted survey of the literature rather than a systematic analysis. We used \textit{Google Scholar} as the primary search tool to identify relevant publications at the intersection of artificial intelligence, machine learning, hydrogen dispersion, and hydrogen bunkering, and subsequently verified the quality and venue of the selected works through \textit{Web of Science}. This approach ensures coverage of the most relevant and recent contributions, but it does not claim exhaustive completeness.

\subsubsection{AI systems for hydrogen dispersion}\label{sec121}
Existing AI-enabled hydrogen dispersion research focuses mainly on land-based refueling stations in open environments \cite{wangPhysicsinformedNeuralNetworks2025, wangRealtimeReconstructionHydrogen2025, hePredictionModelEvolution2024} and relies mostly on synthetic CFD data.

The first model employs a vector-quantized variational autoencoder architecture to reconstruct the concentration field of a hydrogen leak \cite{wangRealtimeReconstructionHydrogen2025}. This approach differs from the required task of forecasting an evolving hydrogen dispersion and its concentration field, as its objective is reconstruction. To make decisions about triggering potential safety barriers or emergency shutdowns, the static reconstruction of concentration fields is not sufficient to support such decision during operation.

The second model functions as a CFD surrogate, using image-to-image translation to accelerate simulation times \cite{hePredictionModelEvolution2024}. This approach reduces the computational expense compared to CFD simulations, and thus its general objective is similar to this work. However, its reliance on dense concentration-field data comparable to the training CFD grids renders its inapplicable to real-world scenarios where such data are typically unavailable due to sparse sensor networks.

The third model introduces a physics-informed deep learning framework to forecast hydrogen concentration within a fixed spatial domain \cite{wangPhysicsinformedNeuralNetworks2025}. However, physics-informed models require explicit initial and boundary conditions for their governing partial differential equations. An assumption that does not reflect the dynamic nature of real-world leakage events and that fails silently when conditions change.

The fourth model uses experimental data, thereby grounding the analysis in real-world physics \cite{alfariziAccidentPreventionLiquid2023}. It frames the problem as classification task, detecting the presence of a leak rather than forecasting the evolving concentration field. The study employs a Random Forest model on individual, temporally dependent samples to predict whether, after 200 seconds, the hydrogen concentration at a given location will exceed the lower flammability limit. The reported accuracy metrics are promising but given the smoothing tendencies of decision tree ensembles its generalization for unseen scenarios is questionable. The proposed approach is useful for simple alarm systems but falls short of providing a forecast of concentration levels that is needed to enhance the situational awareness of operators.

Whereas each of the mentioned studies represents a valuable addition to the study of hydrogen dispersion and how AI-enabled system can be used to improve safe and reliable operation of hydrogen bunkering, all of them have their individual limitations and fall short on three general things:
\begin{enumerate}
    \item They do not explicitly address compressed hydrogen bunkering in confined spaces that is important for the diffusion of hydrogen into the maritime sector.
    \item They propose methods for hydrogen dispersion but they do not discuss the implications on the system-level and how such AI systems can be incorporated into safety-related systems. This is an important stage for actual diffusion of AI systems into real-world applications.
    \item They do not address uncertainties in their model predictions. Especially during runtime, communicating uncertainties in decisions is important for human operators to make informed decisions.
\end{enumerate}

\subsubsection{AI systems in safety-critical systems}\label{sec122}
Before reviewing the literature, it is useful to distinguish two terms. An AI system comprises one or more machine-learning models together with the necessary supporting components, such as input processing, feature extraction, and output processing, required for the system to perform its intended function. An AI-enabled system, by contrast, is the broader engineered system into which one or more AI systems are integrated \cite{detnorskeveritasAssuranceAIenabledSystems2023}. It includes conventional hardware, software, human operators, and safety mechanisms that interact with the AI components. For instance, a compressed hydrogen bunkering station can be an AI-enabled system if it contains an AI system that forecasts leakage dispersion from sensor data, but it also includes the hydrogen sensors, the bunkering control room, and an independent emergency shutdown system that closes manifold valves regardless of the AI output.

The incorporation of AI systems into safety-critical systems has accelerated with growing data accessibility, computational resources, and system complexity that may be difficult to control with traditional methods. Yet concerns about reliability and overall safety make the diffusion non-trivial. In 2024, the International Organization for Standardization (ISO) and the International Electrotechnical Commission (IEC) published TR~5469 \cite{iso/iectr5469:2024ArtificialIntelligenceFunctional2024}, which is meant as recommendation for system developers and classifies AI systems in safety-related systems by usage level and technology class to determine whether existing risk reduction techniques can be applied. For instance, usage level "D" describes the implementation of an AI system in a safety-related system such that existing risk reduction measure are sufficient and the AI system is not part of the safety function. In general, the report surveys lifecycle challenges but remains an initial guide. It offers no standard and leaves gaps without recommendations.

To improve the reliability of the system, \cite{flehmigImplementingArtificialIntelligence2024a} extends the ISO/IEC TR 5469 architectural patterns with a mandatory monitoring component that observes AI inputs, internal processes and outputs, and an input guard component. The monitored information is relayed to a human supervisor who decides whether to fall back to conventional control or retain the AI system in the loop. An updating component, triggered manually by the supervisor, ensures continuous model validity. Complementing these architectural views, \cite{habliBIGArgumentAI2025, wolnyNeedSystematicApproaches2025, dobbeSystemSafetyArtificial2022} argue that assurance of AI-enabled systems must originate from a system-theoretic perspective rather than from the AI system in isolation. While these works provide valuable frameworks, they lack a concrete real-wold use case that demonstrates how these principles can be instantiated in an operational safety-related system.

\subsection{Research gap and problem formulation} \label{sec13}
The literature on AI-enabled hydrogen safety-related systems exhibits two gaps. First, existing AI-based dispersion models target hydrogen refueling for vehicles rather than compressed $H_2$ bunkering in confined spaces for the maritime sector, they rely mostly on CFD data rather than experimental data, and they do not compute uncertainties with their point estimations. No framework exists that combines real-time leakage-rate estimation with probabilistic, short-horizon forecasting of concentration fields using Gaussian processes that enable calibrated uncertainty estimation and supports operator situational awareness through sensor augmentation.

Second, the literature is divided in two fields: publications on AI systems typically stop at the model boundary, whereas publications on system design remain often abstract and lack a guidance on how AI system may be incorporated, especially in safety-related systems. An integrated treatment that delivers both a tested predictive model and a system design for its incorporation is missing.

This paper addresses both gaps. Its contributions are:
\begin{enumerate}
    \item A novel two-stage framework for a safety-related system for compressed $H_2$ bunkering that estimates the unknown leakage rate via an ensemble of multi-layer perceptrons and forecasts the spatiotemporal dispersion field via Gaussian process, thereby enhancing operator situational awareness and enabling preventive safety barriers before an emergency shutdown becomes necessary.
    \item An initial system design for incorporating this AI system into a safety-related system for bunkering in compliance with ISO/IEC TR 5469, and a demonstration that the AI system can be segregated as usage level "D" without compromising the independent emergency shutdown function.
\end{enumerate}

To the best of our knowledge, no open dataset exists for hydrogen dispersion inside a ship's tank connection space. To enable reproducible benchmarking of the AI framework, we therefore demonstrate it on an open dataset of hydrogen dispersion in a confined open-ended rectangular channel \cite{henriksenStudyHydrogenDispersion2025, fossumNumericalSimulationHydrogen2024, mathiasExperimentalDataHydrogen2025}. This dataset captures the key physical phenomena relevant to bunkering safety, i.e., buoyant gas accumulation in a confined volume, and a sparse sensor network for monitoring, while providing boundary conditions necessary for model validation. The framework is build so that the GP kernel structure and sensor-feature extraction can be transferred to other geometries once operational data or validated CFD become available.

\section{Methodology} \label{sec2}
This section is divided into three different methodological steps: system design, dataset acquisition and processing, and the two-stage framework architecture. The \textit{system design} addresses the challenge of an initial system design for incorporating an AI system into a safety-related system such as the compressed $H_2$ bunkering system, \textit{dataset acquisition and processing} introduces the experimental setup which was used to acquire the data and discusses processing steps that where performed. Finally, the \textit{two-stage framework architecture} briefly describes the underlying algorithm of GPs, explains the model selection, the training and testing setup as well as the inference which is how the AI system is used during runtime and describes the two-stage framework.

\subsection{System Design} \label{sec21}
To develop the system design for an AI-enabled safety-related system for compressed $H_2$ bunkering, we followed three consecutive steps. First, we reviewed literature on compressed $H_2$ bunkering systems, their descriptions, and available design indicators. Second, we examined current guidelines and publications on the incorporation of AI systems into safety-related systems. Third, we evaluated literature on safe system design for complex systems.

\subsubsection{Compressed \texorpdfstring{$H_2$}{H2} Bunkering System} \label{sec211}
\begin{figure}
    \centering
    \includegraphics[width=0.7\linewidth]{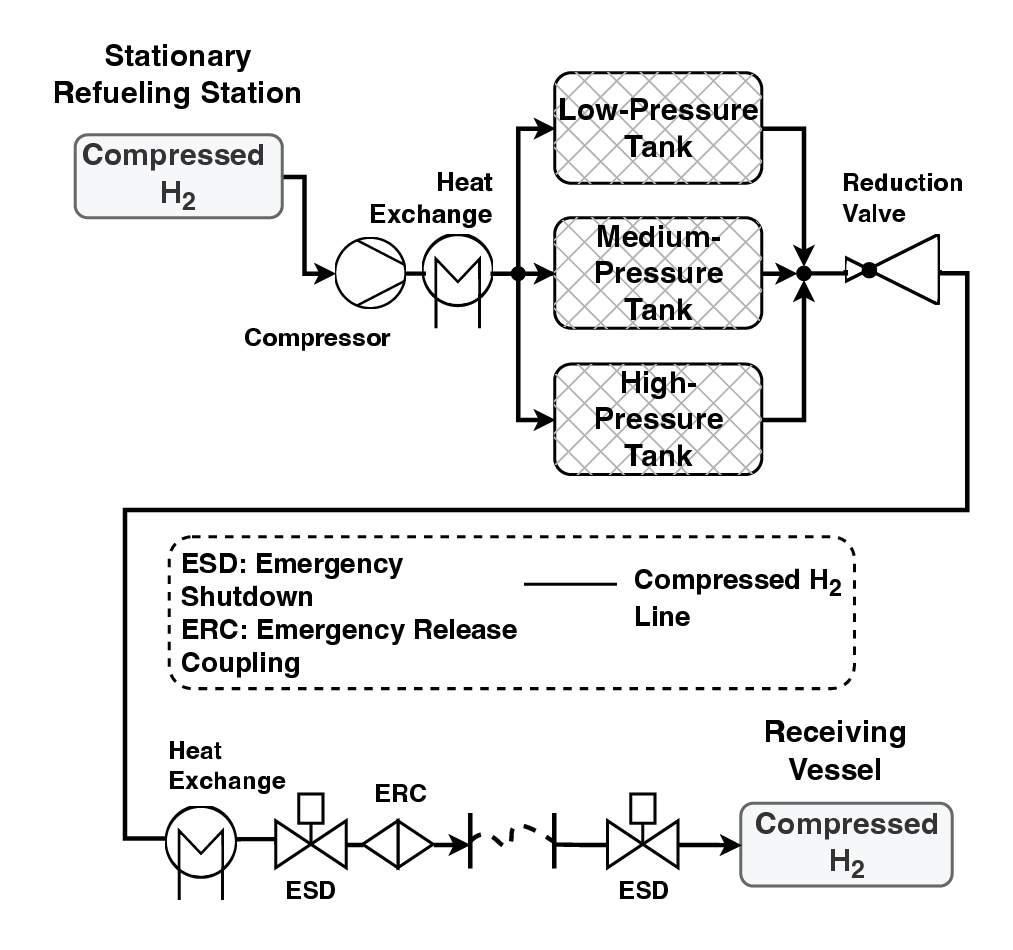}
    \caption{Generic system diagram for compressed $H_2$ bunkering system based on \cite{zhakaHydrogenFuelMaritime2024,tamburiniPerformanceAssessmentSafety2024, europeanmaritimesafetyagencyHazardIdentificationGeneric2024}}
    \label{fig:systemDiag}
\end{figure}
To establish a baseline system design for a compressed $H_2$ bunkering system with an incorporated AI system, we reviewed literature on $H_2$ bunkering to derive a design without any AI system. A report on $H_2$ for vessels including the bunkering process \cite{handbookHydrogenVessels} and a publication on general $H_2$ refueling systems \cite{liuHydrogenRefuelingStations2025} were used as starting point. Further, publications that discuss the adoption of hydrogen as fuel in the maritime sector were analyzed \cite{zhakaHydrogenFuelMaritime2024, zhakaLargescaleHydrogenRefuelling2026}. These publications discuss also the challenges and give recommendations on preventive and mitigation measures.

One main hazard during bunkering is the leakage of hydrogen \cite{liuHydrogenRefuelingStations2025, handbookHydrogenVessels, zhakaHydrogenFuelMaritime2024}. Preventive measures are addressing the design of the piping system as well as the construction and support of the ship bunkering manifold. Possible mitigation measures describe passive measures like bunkering station on unobstructed open deck and minimizing the volume of the bunkering station as well as active measures such as leak detection and shutdown, shutdown valves and emergency shutdown communication system \cite{handbookHydrogenVessels}. The system diagram of a generic compressed $H_2$ bunkering system can be seen in Figure \ref{fig:systemDiag}. The emergency shutdown (ESD) is the main safety system of the bunkering system. Triggering it means to quickly and safely shutdown the bunkering operation by closing the manifold valves, stopping pumps, and closing tank filling valves. The ESD must be capable to activation from both, the bunkering receiving vessel and the stationary refueling station. In this work, we assume a remote bunkering control room where operators monitor the process via CCTV and essential measurements such as tank pressure, tank temperature, hydrogen concentration measurements, and automatic and manual shutdown indicators are also shown at this location \cite{handbookHydrogenVessels}. Whereas there are different causes of an ESD besides hydrogen leakage such as fire detection or power failure, we only consider the hydrogen concentration which exceeds the LFL. For this work, we are especially interested in the leakage of compressed $H_2$ in confined spaces like the tank connection space inside the vessel. The leakage here differs from open-atmosphere releases because it is in a closed room where wind effects are negligible. Typically, the released hydrogen mixture mixes with the room atmosphere and disperses outwards through venting holes with active or passive ventilation.

\subsubsection{Incorporation of AI systems into safety-critical systems}\label{sec212}
The incorporation of AI systems into safety-critical systems is not straightforward due to severe consequences of system failures. As of the writing of this work, there are no existing standards on this issue but guidelines from ISO and IEC \cite{iso/iectr5469:2024ArtificialIntelligenceFunctional2024} with a standard under development \textit{ISO/IEC CD TS 22440}. This work will use the technical report and an extended view on the architectural patterns and framework for operation from \cite{flehmigImplementingArtificialIntelligence2024a} to guide the system design and aim to comply with the provided guidelines. For safety-critical systems, a safety-related system is needed. It consists of one or several safety functions and other risk reduction measures to maintain the safe state of the system. 

The first task for the system design of an AI-enabled safety-related system is the classification of the AI system by its usage level. The usage levels are defined by six different levels from "A1" describing an AI systems which is used in a safety system and which conducts automated decision-making of system functions to level "D" which describes AI systems that are not part of a safety function and has no impact on the safety function due to sufficient segregation and behavior control. For instance, a conventional safety logic controller is in control of an emergency shutdown system as the safety function while an AI system independent from this provides additional information about the system's state. After determining the usage level, the AI technology class is reviewed which basically determines if existing functional safety international standards can be used to reduce risk. Class III describes AI systems that cannot be developed and reviewed using existing techniques, class II AI systems cannot be fully developed and reviewed, while class I can be. Based on this scheme, AI systems which are classified as usage level "D" can be implemented because existing risk reduction measures can be applied. For instance, ISO 61508 provides such risk reduction measures. This usage level also allows models for online updating which can be helpful in complex environments. Our system design incorporates the AI systems as usage level "D" where it is part of the safety-related system by enhancing the situational awareness of operators but not part of the safety function as this is an independent emergency shutdown system based on a conventional safety logic solver. The input for the AI system and the safety function can be the same as it is the case in our system but does not necessarily be the case. For instance, in systems where the AI system is performing some kind of image segmentation conventional controller in the safety function might not be able to handle this type of unstructured data in a reliable way.

To tolerate errors without loss of reliability properties, a robust architecture around the AI system is needed. Here, the technical report suggests three different architectural patterns that can co-exist or used as alternatives to each other:
\begin{itemize}
    \item Backup function to the machine learning (ML) component is designed with non-AI techniques. This backup decision system is for example an alternative controller of fail-safe null action. The backup action allows the use of detection methods to switch the output when unsafe conditions are detected.
    \item A safe subset of the action space is determined using a supervisor function with constraints or limits.
    \item ML redundancy with output voters or aggregators is considered.
\end{itemize}
    
\begin{figure}
    \centering
    \includegraphics[width=0.9\linewidth]{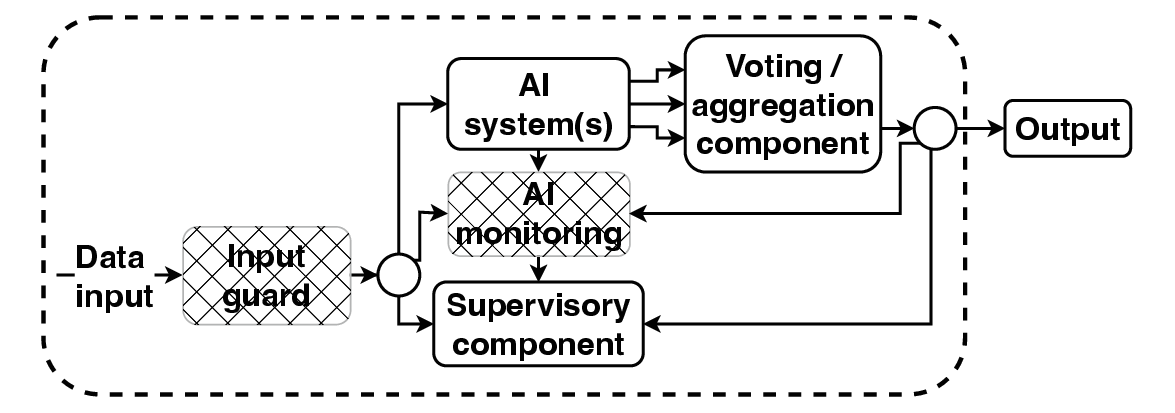}
    \caption{Generic architectural pattern for AI system(s) in safety-related systems based on \cite{flehmigImplementingArtificialIntelligence2024a}. Raw data inputs first pass through an input guard that validates format, range, and integrity. The data then feed the AI system(s), a monitoring component that observes both inputs and internal AI processes, and a supervisory component representing the human operator. The AI outputs are aggregated by a voting component before being relayed to the downstream process. The monitoring component informs the supervisor, who retains authority to override the output of the AI system(s).}
    \label{fig:architecturalPattern}
\end{figure}

We chose to use a combination of all three suggested architectural patterns with a supervisory component that represents human operators in the bunkering control room and an independent monitoring component for the AI system as proposed in \cite{flehmigImplementingArtificialIntelligence2024a}. The monitoring component can be implemented with several layers to detect inputs that are out of the learned distribution and to identify unusual internal operations \cite{flehmigPerspectivesReliabilityMonitoring2026}. An input guard component is also implemented to provide an online validation of the data stream. This is performed continuous and in real time by checking signal integrity. For more in-depth data analysis, an offline procedure in regular intervals is needed. The option for a safety backup system is not needed in this work because the AI system is not implemented as safety function but independent from it. This independence from the safety function also ensures that faults by the input guard or the AI monitoring are not affecting the overall safety of the system. Faults in either of these components may trigger unnecessary action but do not comprise the safety of the system.

Relevant but also excluded in this work due to the scope is the updating component to ensure reliability during runtime despite changes in the data distribution. The generic architectural pattern, we used in our work is displayed in Figure~\ref{fig:architecturalPattern}.

The training of AI systems is inherently probabilistic and trained AI systems are often non-deterministic or they are deterministic but due to complex environment their output seems to be non-deterministic. This makes it arguable impossible to guarantee well generalization on unseen operational conditions. In addition, AI systems are self-organized rather than human-programmed, i.e., their rules are not explicitly programmed but learned from finite data which makes it difficult to understand system behavior and identify faults. To address these challenges, the ISO/IEC TR 5469 \cite{iso/iectr5469:2024ArtificialIntelligenceFunctional2024} gives a perspective on using the "proven-in-use" concept from software safety standards \cite{zotero-item-763} and suggests an adaption for AI systems. Proven-in-use applies only to specific products, they must be clearly restricted in functionalities, and the safety integrity level of the safety-related system that incorporates these elements remains unaffected. Evidence is drawn from documented operational evidence, specifically:

\begin{itemize}
    \item \textbf{Conditions of use:} documented operational envelope including modes of use, configuration, and human factors that could trigger systematic faults.
    \item \textbf{Failure-rate history:} demonstrated dangerous failure rate below the required bound, supported by an effective reporting system.
    \item \textbf{Impact analysis:} analytical and test-based demonstration that differences between previous and anticipated conditions of use do not introduce dangerous systematic faults.
\end{itemize}

Proven-in-use is well defined for conventional software, but its direct adoption to AI systems remains an open question. Therefore, we draw on the general IEC~61508 principles of systematic fault avoidance and control, translated to AI training, testing, and validation, and emphasize runtime monitoring. Our discussion offers a perspective on how these functional safety principles might be extended to AI-enabled systems, rather than claiming a complete solution.

\subsubsection{Safe System Design}\label{sec213}
Safety must be designed into a system, just identifying and assessing hazards is not enough to make a system safe. Therefore, addressing safety early in the design process of a system can improve the safety of the system while saving cost by avoiding expensive retro-fitting or changes in the design late in the development process. The safety in a system is not just influenced by physical hazards but also by complex interactions. The introduction of software and more recently AI systems has increased the complexity of systems. The challenge with AI systems is similar to software applications where hazards can stem from unsafe interactions of subsystems and unpredictable behavior. Thus, when we incorporate AI systems into safety-related systems the detection and handling of unsafe and unreliable system behavior must be considered. On the other hand, the incorporation of AI systems also introduces new possibilities for improving system safety and reliability in terms of increased functionality through advanced data analysis and more powerful protection mechanisms \cite{levesonIntroductionSystemSafety2023, levesonDesignAssuranceControl2025}.

A common technique for hazard analysis in complex systems, that can be used to design a safe system is called system-theoretic process analysis (STPA). STPA is a hazard analysis technique designed to address system complexity by using a system-theoretic approach and treating safety as a control problem. It recognizes that accidents can arise not only from component failure, but from unsafe or missing control actions within complex interactions \citep{rausandRiskAssessmentTheory2020}. STPA complements traditional hazard analysis techniques by capturing the increased complexity that AI systems introduce through complex interaction between technical systems, humans, and operational processes.

\subsection{Dataset acquisition and processing}\label{sec22}
As a proof-of-concept for confined-space hydrogen dispersion forecasting, we used the open dataset on hydrogen dispersion from \cite{mathiasExperimentalDataHydrogen2025}. They investigated the downward release and dispersion of hydrogen. One of the main goals of this experimental campaign was to generate a broad range of quantitative data on large-scale hydrogen dispersion that was well-suited for validating CFD models. Additionally, we had access to the data from the corresponding CFD model which got validated on this experimental data and was discussed here \cite{fossumNumericalSimulationHydrogen2024}. In total, we had data from 32 tests with 22 experimental tests and 10 CFD tests. 26 tests are unique because 6 of the CFD tests were used to validate the model against the experimental data and thus are approximately duplications but still valid to use.

\subsubsection{Experimental setup}\label{sec221}
The experimental data investigates the downward release and dispersion of hydrogen in a 5.8\,m long open-ended rectangular channel. Hydrogen was released at the top of the channel, 0.5\,m from the closed end, with mass flow rates ranging from 0.04 to 1.27\,g/s. 29 hydrogen concentration sensors (XEN-5320 Xensors) were mounted inside the channel with an accuracy of $\pm$1\%. In a subset of experiments a wooden obstruction was placed 2.1\,m from the closed end to perturb the flow. Release durations varied between 60 and 240 seconds, chosen according to the time required to reach steady-state concentrations. A comprehensive description of the channel geometry, instrumentation, and test matrix is given in \cite{henriksenStudyHydrogenDispersion2025}. Whereas the geometry is a simplified channel rather than a ship's tank connection space, it preserves the safety-critical physics of interest: a confined, poorly ventilated volume with buoyant hydrogen release.
\begin{figure}
    \centering
    \includegraphics[width=0.9\linewidth]{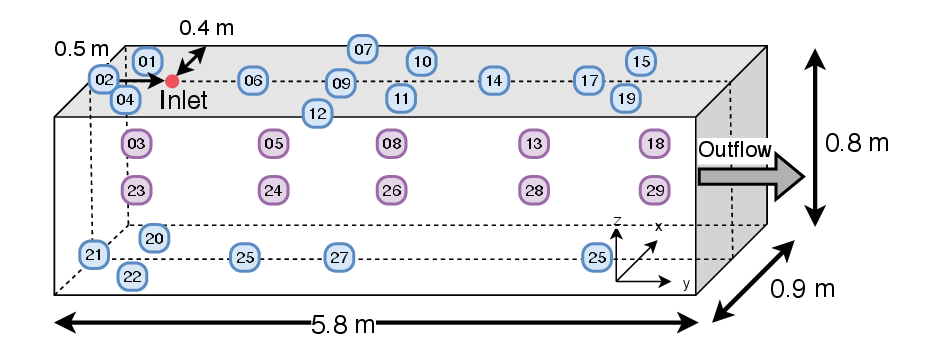}
    \caption{The placement of the hydrogen sensors in the experimental setup based on \cite{fossumNumericalSimulationHydrogen2024}.}
    \label{fig:SensorPlace}
\end{figure}
The corresponding CFD data were generated with OpenFOAM version 10 to replicate the experimental conditions described above. \cite{fossumNumericalSimulationHydrogen2024} provide the full details on the computational domain, mesh, governing equations, and numerical setup.

\subsubsection{Data processing for training}\label{sec222}
The complete dataset comprises 22 experimental tests and 10 CFD tests. Each sample contains time $t$, leakage rate $\hat{m}$, sensor coordinates $(x,y,z)$, and measured $H_2$ concentration $c$. The CFD data were originally sampled at 100\,Hz while the experimental data have an effective frequency of approximately 1.6\,Hz. The CFD series was therefore down-sampled to match the experimental frequency. Because the experimental recordings contain an initial idling period before hydrogen enters the channel and a trailing segment after the release is stopped, each test was trimmed to retain only the exposed leakage interval. After trimming, the combined dataset contains approximately 182\,000 samples.
\begin{figure}
    \centering
    \includegraphics[width=0.9\linewidth]{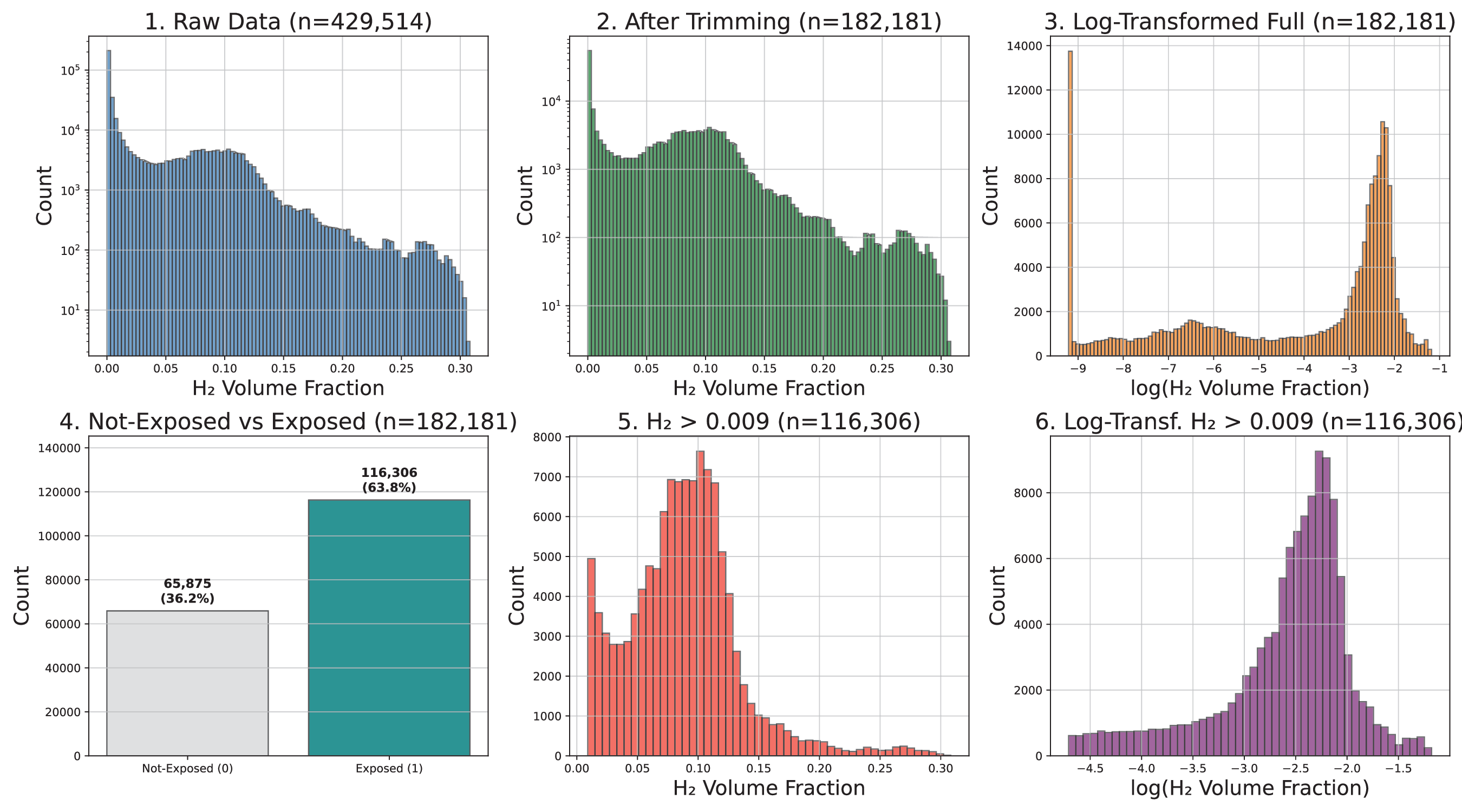}
    \caption{Data processing steps for the hydrogen concentration dataset with their practical safety implications. \textbf{Top row:} (1) Raw data ($n=429\,514$) showing extreme right skew. (2) Trimmed data ($n=182\,181$) after removing idle periods before and after the exposed leakage interval; transformation step to remove non-leakage related data that would introduce a bias. (3) Log-transformed trimmed data revealing a bimodal structure with distinct not-exposed (near-zero) and exposed sensor populations; transformation reveals high number of not-exposed sensors that should be excluded from the dispersion forecast. \textbf{Bottom row:} (4) Not-exposed vs. exposed split using a threshold of 0.009 H$_2$ volume fraction, consistent with the sensor accuracy of $\pm$1\%. (5) Distribution of the exposed sensors before transformation; represents relevant sensors for the forecast. (6) Log-transformed exposed subset showing an approximately Gaussian distribution; transformation enables Gaussian likelihood within the \textit{GPyTorch} framework but comes with information loss.}
    \label{fig:AllHisto}
\end{figure}
A log transformation was initially applied to the trimmed data to mitigate the right skew. As shown in Figure~\ref{fig:AllHisto}~(top right), the log-transformed trimmed distribution has almost two modes: one mode corresponds to sensors that are effectively not exposed with concentrations near zero, and a second mode corresponds to sensors that have detected hydrogen. Because XEN-5320 sensors have an accuracy of $\pm$1\% full scale, a threshold of 0.009 was chosen to separate not-exposed from exposed readings. Values below this threshold are treated as background noise. The threshold was chosen slightly below the sensor's accuracy so that the very first indications of a leak are captured.

For the exposed subset, the $H_2$ concentration distribution, Figure~\ref{fig:AllHisto}~(bottom middle), remains right skewed. Applying a log transformation to this exposed-only subset produces an approximately Gaussian distribution, Figure~\ref{fig:AllHisto}~(bottom right).

For each test, the leakage onset is defined as the first time instant at which any sensor in the array exceeds the 0.009 threshold. From the 10-second window immediately following this onset, a set of 15 features is extracted for the first-stage Multi-layer Perceptron (MLP) ensemble. At each time step within this interval, a set of 10 snapshot features is computed: number of exposed sensors, elapsed time since onset, maximum concentration, mean concentration, standard deviation of concentration, spatial spread in the $y$-direction, spatial spread in the $z$-direction, and the weighted spatial spread in $y$ and $z$. In addition, five temporal derivatives computed over three consecutive time steps: rate of change of exposed sensors, rate of change of maximum concentration, rate of change of mean concentration, rate of change of spatial change in $z$-direction, rate of change of weighted spread in $z$-direction. This yields a 15-dimensional feature vector at each time step that captures both the current spatial state of the dispersion and its short-term evolution during the critical early phase of the leak.

\begin{table}[htbp]
\centering
\caption{Descriptive statistics of the trimmed dataset ($n=182\,181$).}
\label{tab:statistics}
\begin{tabular}{@{}lrrrr@{}}
\toprule
Feature & Mean & Std & Min & Max \\
\midrule
time $t$ (s) & 74.28 & 55.98 & 0.00 & 239.95 \\
leakage rate $\hat{m}$ (g/s) & 0.35 & 0.32 & 0.03 & 1.27 \\
$x$ (m) & 0.46 & 0.20 & 0.00 & 0.90 \\
$y$ (m) & 2.17 & 1.61 & 0.00 & 4.68 \\
$z$ (m) & 0.49 & 0.31 & 0.00 & 0.80 \\
H$_2$ conc.\ $c$ (\%) & 0.06 & 0.06 & 0.00 & 0.31 \\
\bottomrule
\end{tabular}
\end{table}
Consequently, two processed datasets are retained for the modeling:
\begin{enumerate}
    \item \textbf{Trimmed dataset} ($n \approx 182\,000$): all trimmed samples with the binary exposed/not-exposed label derived from the 0.009 threshold. This dataset supports the first-stage MLP ensemble and auxiliary analyses of sensor coverage. The statistics are shown in Tab.~\ref{tab:statistics}.
    \item \textbf{Exposed-only dataset}: samples above the 0.009 threshold with and without log-transformed $H_2$ concentration as the regression target.
\end{enumerate}

Finally, all negative concentration values arising from measurement noise were clipped.

\subsection{Framework architecture}\label{sec23}
The proposed framework consists of two sequential models. Stage~1 is a MLP ensemble that estimates the leakage rate from features extracted during the first 10 seconds after the leakage onset. Stage~2 is a regressor that predicts the concentration field using the estimated leakage rate.

\subsubsection{Gaussian processes}
Gaussian Processes (GPs) are a simple and general class of models of functions. They offer powerful non-parametric function priors that are useful in a variety of regression and classification tasks. For a given input space $R^d$ GPs are entirely specified by a covariance function or kernel $k(\mathbf{x},\mathbf{x}')$ and a mean function $m(\mathbf{x})$:
\begin{equation}
m(\mathbf{x}) = \mathbb{E}\!\left[f(\mathbf{x})\right],
\end{equation}

\begin{equation}
k(\mathbf{x}, \mathbf{x}') = 
\mathbb{E}\!\left[\left(f(\mathbf{x}) - \mu(\mathbf{x})\right)
\left(f(\mathbf{x}') - \mu(\mathbf{x}')\right)\right].
\end{equation}

The GP is written as

\begin{equation}
    f(x) \sim \mathcal{GP}\bigl(m(\mathbf{x}),k(\mathbf{x},\mathbf{x}')\bigr)
\end{equation}

Different choice of $m$ and $k$ allow prior information about the generative process to be encoded \cite{rasmussenGaussianProcessesMachine2006, duvenaud2014automatic}. The kind of structure a GP model can capture is entirely defined by its kernel. This also describes the main difficulty of GPs: constructing an appropriate kernel. Some properties of GPs can be seen as especially well-suited for the safety-critical domain:
\begin{itemize}
    \item \textbf{Expressivity and easy to analyze:} Through the choice of covariance function, we can express a wide range of modeling assumptions. It also gives a set of computable functions which gives us rather simple models that can be used as well-understood building blocks.
    \item \textbf{No overfitting:} GPs are less prone to overfitting in comparison to neural networks for instance.
    \item \textbf{Uncertainty:} GPs naturally compute uncertainties of their predictions.
\end{itemize}

To express complex structure in data, kernels can be combined by addition or multiplication. For more details on this, we recommend \cite{duvenaud2014automatic}. The second challenge is the slow inference which limits the amount of data that can be used during training. Making exact predictions with GPs is $O(N^3)$, which can be prohibitive for large $N$. Approximate inference techniques, discussed in Section~\ref{sec232}, overcome this issue to some extent, yet computational cost remains a limiting factor for datasets exceeding one million points.

\subsubsection{Training data}\label{sec231}
The two processed datasets from Section~\ref{sec222} are used for model development. The trimmed dataset contains all samples from the leakage intervals with the features: time $t$, leakage rate $\hat{m}$, Cartesian coordinates $(x,y,z)$ of each sensor, and measured $H_2$ concentration $c$. The exposed-only dataset contains the same features but retains only the exposed samples, with the $H_2$ concentration to serve as the regression target.

Both datasets were split into training and testing sets by scenario to ensure that all time steps belonging to one experiment or CFD test reside entirely in one split. This prevents data leakage across the temporal dimension and preserves the physical coherence of each leakage event. The train/test ratio was varied between approximately 80/20 and 70/30. The training portion was further subdivided into training and validation sets following standard machine-learning practice, see Table~\ref{tab:dataset_split}. The validation data were used for hyperparameter tuning and early stopping, while the test set remained unseen until final evaluation.

\subsubsection{Model architectures}\label{sec232}
For Stage~1, an ensemble of MLPs is trained to map 29 sensor measurements over three consecutive time steps concatenated with a 15-dimensional feature vector to the scalar leakage rate estimate $\hat{m}$. The features are extracted at each time step within the 10-second window immediately following leakage onset, defined as the first time any sensor exceeds the 0.009 $H_2$ volume fraction threshold. The feature vector comprises 10 snapshot quantities and 5 temporal features. The MLP ensemble consists of multiple independently trained networks with different random initializations. The mean of the individual predictions yields the leakage rate estimate $\hat{m}$, whereas the variance across ensemble members provides a heuristic uncertainty that can flag unreliable estimates. The MLPs are implemented and trained in \textit{PyTorch} using standard fully-connected layers with ReLU activation and dropout for regularization.

For Stage~2, a GP regression model is trained on the exposed-only dataset. It maps the input vector \newline $\mathbf{x} = [t, \hat{m}, x, y, z]^\top$, time $t$, leakage rate $\hat{m}$, and sensor coordinates$(x,y,z)$, to the $H_2$ concentration $c$ using an additive kernel structure. The posterior mean $m_{\text{GP}}$ and covariance $k_{\text{GP}}$ provide the predicted concentration field and its uncertainty.

Manual kernel construction is difficult, so we adopted an additive decomposition method \cite{duvenaudAdditiveGaussianProcesses2011}. This method provides a fast and simple way to discover a suitable structure. Instead of building a kernel structure piece-by-piece, which can be difficult and time-consuming, additive GPs start with many candidate structures and adjust kernel parameters to discard those not supported by the data. We assign an independent radial-basis function (RBF) kernel $k_d$ to each of the five input dimensions. The full kernel is the sum of all $2^5-1 = 31$ products of these base kernels, grouped by interaction order:

\begin{align}
k_{\text{add}}^{(1)}(\mathbf{x},\mathbf{x}') &= 
\theta_{0}k_{0}(t,t') 
+ \theta_{1}k_{1}(\hat{m},\hat{m}') 
+ \theta_{2}k_{2}(x,x') \nonumber\\
&\quad
+ \theta_{3}k_{3}(y,y') 
+ \theta_{4}k_{4}(z,z'),
\label{eq:first_order}
\\[4pt]
k_{\text{add}}^{(2)}(\mathbf{x},\mathbf{x}') &= 
\theta_{0,1}k_{0}k_{1}
+ \theta_{0,2}k_{0}k_{2}
+ \theta_{0,3}k_{0}k_{3} \nonumber\\
&\quad 
+ \theta_{0,4}k_{0}k_{4}
+ \theta_{1,2}k_{1}k_{2}
+ \theta_{1,3}k_{1}k_{3} \nonumber\\
&\quad 
+ \theta_{1,4}k_{1}k_{4}
+ \theta_{2,3}k_{2}k_{3}
+ \theta_{2,4}k_{2}k_{4} \nonumber\\
&\quad 
+ \theta_{3,4}k_{3}k_{4},
\label{eq:second_order}
\\[4pt]
k_{\text{add}}(\mathbf{x},\mathbf{x}') &= 
k_{\text{add}}^{(1)}(\mathbf{x},\mathbf{x}') 
+ k_{\text{add}}^{(2)}(\mathbf{x},\mathbf{x}') \nonumber\\
&\quad 
+ \sum_{\substack{\mathcal{S}\subseteq\{0,\dots,4\}\\|\mathcal{S}|\ge 3}} 
\theta_{\mathcal{S}}\!\prod_{d\in\mathcal{S}}\!k_{d}(x_{d},x_{d}').
\label{eq:general_additive}
\end{align}

where $\mathcal{S}$ indexes every non-empty subset of the five dimensions. The terms can be grouped by interaction order. Unlike the original formulation of \cite{duvenaudAdditiveGaussianProcesses2011}, which shares a single output scale $\theta$ across all terms of the same order, we adjusted the method by assigning a learnable output scale to every subset. This yields $2^D-1$ scaling parameters rather than $D$, increasing the model capacity while preserving interpretability. When a subset output scale collapses toward zero, the corresponding interaction is effectively discarded by the data. The learned output scales provided a direct measure of term importance, which we used to reduce kernel complexity by manually discarding the least influential terms after initial training. A new additive kernel was then constructed from the 12 most important combinations identified in this analysis. Each retained term possesses its own output scale, while the lengthscales of its constituent dimensions remain independently parameterized.

The kernels are implemented in \textit{GPyTorch} \cite{gardner2018gpytorch}. The forward pass iterates over all dimension subsets, computes the product of the active base kernels, and scales the result by the corresponding learnable parameter.

Exact GP inference is $\mathcal{O}(N^3)$ and thus infeasible for our dataset of $N \approx 182\,000$ and 5 dimensions regarding our available compute. We therefore employed two different approximation methods. Approximation methods are also necessary for the use of non-Gaussian likelihoods. The Variational Nearest Neighbor (VNNGP) method was introduced by \cite{wuVariationalNearestNeighbor2022} and it uses a sparse approximation of the prior precision matrix. Sparse precision approximation assumes that observations are conditionally independent given their nearest neighboring observations. Such sparse precision approximations are common in spatiotemporal modeling and through the integration in GPs, they make it possible to scale $M$ well beyond the limits of exact GP inference. The advantages of VNNGP are most pronounced on spatiotemporal datasets \cite{wuVariationalNearestNeighbor2022}. Additionally, we employed the Stochastic Variational Gaussian Process (SVGP) method introduced by \cite{hensmanScalableVariationalGaussian2014}. It defines a small set of inducing points $M$. Their positions are typically chosen by k-means algorithm or randomly and can be learned. SVGP assumes that observations are explained through a small number of global latent variables. The assumption is that the full-rank matrix $K_{x,x}$ of the kernel, which would incur a $\mathcal{O}(N^3)$ computation with $N$ being the total observations, can be approximated with a low-rank matrix to reduce the size of any matrix inversion to $M$. This reduces the computational complexity to $\mathcal{O}(M^3)$ with $M \ll N$ \cite{hensmanScalableVariationalGaussian2014}.

For Stage~2, both approximation methods were tested and compared. VNNGP is supposed to perform better on spatiotemporal data where data can vary rapidly with small changes in space and time. This assumption usually holds for geostatistics but not necessarily for the data in this work. On the other hand, SVGP is a popular and well established method for variational approximations to GPs that performs comparable well to VNNGP across different datasets \cite{wuVariationalNearestNeighbor2022}. For the regression task, both Gaussian likelihood and Beta likelihood were tested and compared. The Beta likelihood works naturally for regression tasks on percentage that is bounded in $(0,1)$. Whereas the Gaussian likelihood is the default choice for GPs but for our data it requires a log-transformed hydrogen concentration distribution as shown in Figure~\ref{fig:AllHisto} (6) to perform well.

\subsubsection{Training}\label{sec233}
The MLP ensemble was trained using \newline leave-one-experiment-out cross-validation. In each fold, all samples belonging to a single experiment or CFD test are held out for validation while the remaining experiments are used for training. AdamW \cite{loshchilovDecoupledWeightDecay2017} was used as optimizer for 300 epochs. Dropout regularization with a rate of 0.2 was applied during training. The ensemble comprises multiple networks initialized with different random seeds. The models are trained independently and their predictions are aggregated only at inference.

The SVGP method and the VNNGP method used the exposed-only dataset. The number of training epochs was initially varied between 100 and 300 to explore different training configurations and their impact on the performance. Early stopping on the validation mean absolute error (MAE) was employed to prevent overfitting, so the effective epoch count for each final model was determined by convergence rather than a fixed limit. The optimizer was also AdamW \cite{loshchilovDecoupledWeightDecay2017}. The base learning rate was treated as a hyperparameter and varied across experiments typically ranging from 0.005 to 0.1. The final value for each reported model is given with the corresponding results. The VNNGP method relies on a nearest-neighbor structure and was trained on mini-batches. The SVGP method uses inducing points as representation of the whole data. These inducing points are learned and are initially chosen by using k-means algorithm.

The number of neighbors $k$ was varied between \newline $\{16,32, 64, 128, 256\}$. The number of inducing points was varied between $\{4000, 5000, 6000\}$. For both methods, the choice represents a trade-off between approximation accuracy and computational cost.

All training runs were performed on a single NVIDIA GeForce RTX3090 with 24\,GB of memory. Training times ranged from a few minutes for smaller VNNGP configurations to several days for the largest SVGP models, reflecting the combined effect of approximation type, epoch count, batch size, and learning rate.

\subsubsection{Real-time inference}\label{sec234}

During a bunkering operation the leakage rate of compressed hydrogen into the tank connection space is unknown. The runtime inference therefore proceeds in two consecutive stages: (i) estimation of leakage rate from early-time sensor data via the MLP ensemble, and (ii) short-horizon hydrogen dispersion forecasting via the GP regression model.

From the leakage onset, a 10-second observation window is opened. At each time step within this window, the 15 features described in Section \ref{sec222} are computed. This feature vector is fed to each member of the MLP ensemble concatenated with 29 sensor measurements over three consecutive time steps. After the 10-second window closes, the leakage rate estimation is frozen to the final value $\hat{m}$ obtained at the end of the window and held constant for the remainder of the leak event, reflecting the approximately constant nature of the release.

With the estimated leakage rate $\hat{m}$, the GP regression model is queried at future times $t+h\Delta t$ ($h=1,\dots,H$) and at any spatial coordinate $(x,y,z)$ inside the compartment even where no sensors are placed. Let $\mathbf{x}^{*}=(t+h\Delta t,\hat{m},x,y,z)$. Then
\begin{equation}
c(t+k\Delta t,x,y,z)\sim
\mathcal{GP}\bigl(m_{GP}(\mathbf{x}^{*}),\,
k_{GP}(\mathbf{x}^{*})\bigr).
\label{eq:forecast}
\end{equation}

The mean $m_{GP}$ gives the most likely concentration field. The covariance $k_{GP}$ quantifies the model uncertainty and is displayed to the operator as a confidence map. The total inference latency is dominated by the GP covariance computation, which executes in seconds for a sparse network ($N_s \ll 100$) and a moderate forecast horizon ($H \le 20$).

The two-stage inference, i.e., onset detection, MLP ensemble leakage rate estimation, and GP regression forecasting, is implemented as shown in Algorithm~\ref{alg:runtime} and the corresponding flowchart (Figure~\ref{fig:GP_EKF_flow}, Appendix~A).

\begin{algorithm}[tbp]
\small
\caption{Online H$_2$ dispersion forecasting}
\label{alg:runtime}
\begin{algorithmic}[1]
\Require MLP ensemble $\{\text{MLP}_j\}_{j=1}^{E}$; 
         GP model $\mathcal{GP}_{\text{reg}}$; 
         sensor locations $\{(x_i,y_i,z_i)\}_{i=1}^{N_s}$; 
         threshold $\tau=0.009$; onset $T_{\text{onset}}=10$\,s; 
         horizon $H$
\Ensure Estimates $\{\hat{c}_{t+h},\sigma_{t+h}\}_{h=1}^{H}$
\State $t \gets 1$, $\hat{m} \gets \text{undefined}$
\While{bunkering active}
    \State Acquire readings $\{c_i(t)\}_{i=1}^{N_s}$
    \If{$\max_i c_i(t) > \tau$ \textbf{and} $t \le T_{\text{onset}}$}
        \State \textbf{Stage 1:} Extract 15 features from current 
               and 3 prior time steps
        \State $\hat{m} \gets \frac{1}{E}\sum_{j=1}^{E} 
               \text{MLP}_j(\mathbf{f}_t)$ 
               \Comment{Update estimate}
    \EndIf
    \If{$\hat{m}$ is defined}
        \State \textbf{Stage 2:} Query $\mathcal{GP}_{\text{reg}}$ at 
               $(t+h\Delta t,\hat{m},x,y,z)$
        \Statex \hspace{\algorithmicindent} for $h=1,\dots,H$ 
                 and coords $(x,y,z)$
        \State Store $\{\hat{c}_{t+h}, k_{t+h}\}_{h=1}^{H}$
        \State Display $\hat{c}_{t+1:t+H}$ and $k_{t+1:t+H}$ to operator
    \EndIf
    \State $t \gets t+1$
\EndWhile
\end{algorithmic}
\end{algorithm}

\subsubsection{Evaluation metrics}\label{sec235}
For training the GP models, we maximize the predictive log-likelihood, which is closely related to the variational evidence lower bound but places the logarithm outside the expectation:
\begin{equation}
\small
\begin{split}
    \mathcal{L}_{\text{ELBO}} &= \mathbb{E}_{p_{\text{data}}(y,x)}\!\big[\log p(y|x)\big]
    -\beta\,\text{KL}\!\big[q(u)\|p(u)\big] \\
    &\approx \sum_{i=1}^{N}\log \mathbb{E}_{q(u)}\!\left[\int\! p(y_i|f_i)p(f_i|u,x_i)\,\mathrm{d}f_i\right] \\
    &\qquad -\beta\,\text{KL}\!\big[q(u)\|p(u)\big]
\end{split}
\end{equation}
where $N$ is the total number of datapoints, $q(u)$ is the variational distribution for the inducing function values, $p(u)$ is the prior distribution for the inducing function values, and $\beta$ is a scaling constant that reduces the regularization effect of the KL divergence. This objective function produces better predictive variances than the standard variational ELBO \cite{jankowiakParametricGaussianProcess2019} and was therefore chosen as the training criterion for the regressor.

For validation during hyperparameter tuning and for final testing of the MLP, we report the root mean squared error (RMSE) and mean absolute error (MAE).

\begin{equation}
   \text{RMSE}_{m} = \sqrt{\frac{1}{N_{\text{test}}}
\sum_{i=1}^{N_{\text{test}}} \left(\hat{m}'_i - \hat{m}_i\right)^2} 
\end{equation}

\begin{equation}
\text{MAE}_{m} = \frac{1}{N_{\text{test}}}
\sum_{i=1}^{N_{\text{test}}} \left|\hat{m}'_i - \hat{m}_i\right|
\label{eq:mlp_metrics}
\end{equation}

where $\hat{m}'_i$ is the true mass flow and $\hat{m}_i$ is the ensemble mean prediction. These metrics quantify the accuracy of the critical first-stage estimate that drives the subsequent GP forecast. A large leakage estimation error will propagate directly into the dispersion prediction. In addition, the ensemble variance is monitored as a heuristic uncertainty indicator. High variance signals that the MLPs disagree.

For the dispersion regressor, we report also RMSE and MAE on the validation set. RMSE penalizes large deviations heavily and thus highlights cases where the model misses peak concentrations, which can be critical for safety. MAE provides an interpretable measure of average prediction error in the original units of hydrogen volume fraction.

\section{Results}\label{sec3}
In this section, we present the results of this work which is split into two sections. First, we present a perspective on a possible system design for an AI-enabled safety-related system for $H_2$ bunkering based on the reviewed literature and outline a potential real-world implementation regarding the proven-in-use approach. In the second part of the results, we show the performance of the trained and tested two-stage framework. The results are based on the complete dataset as discussed in Section~\ref{sec22}.

\subsection{System design}\label{sec31}
\begin{figure}
    \centering
    \includegraphics[width=0.8\linewidth]{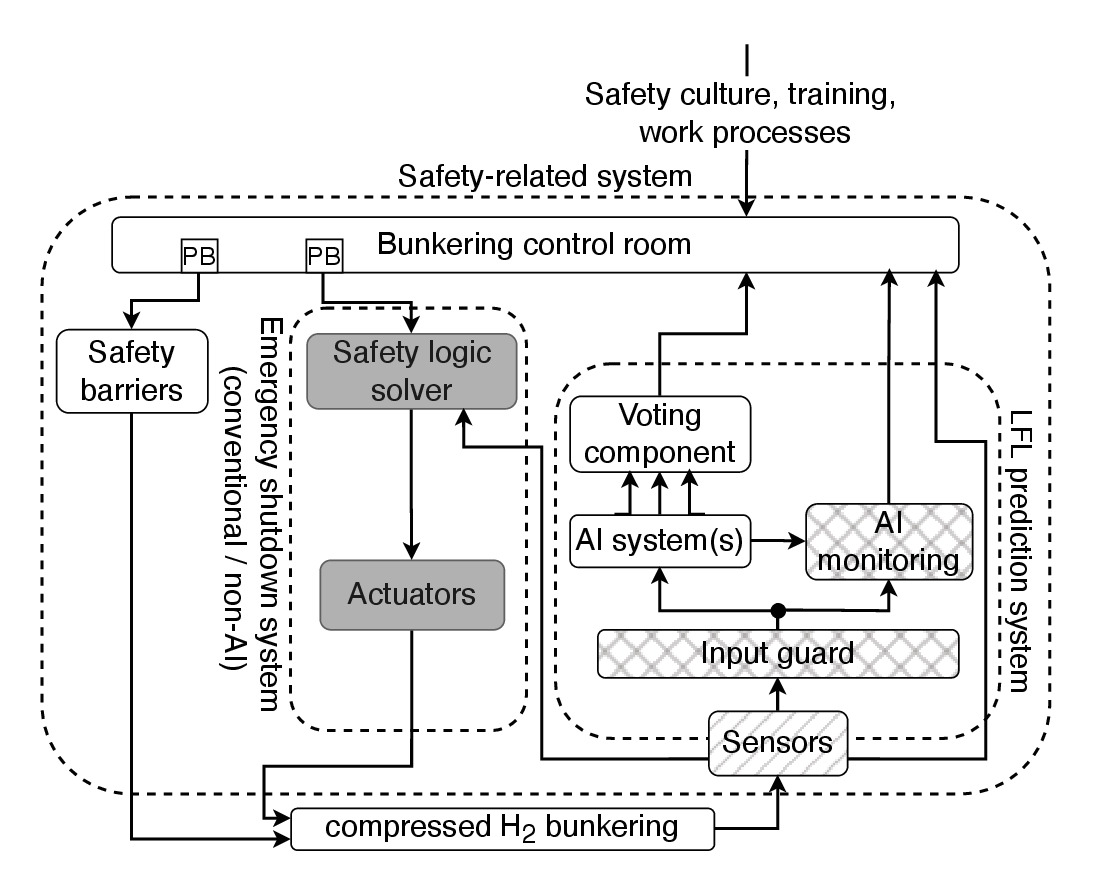}
    \caption{Control structure and system design of the safety-related system within a bunkering system for compressed $H_2$. Filled boxes represent the ESD system that is independent and the safety function of the system. Cross-hatched boxes represent component that are introduced to enhance reliability of the AI system based on \cite{flehmigImplementingArtificialIntelligence2024a}. The sensors are part of the ESD system and their readings are replicated for the use of the AI system.}
    \label{fig:robustImpl}
\end{figure}
The proposed system design is shown in Figure~\ref{fig:robustImpl}. The intrinsic safety of the bunkering system is guaranteed by an independent emergency shutdown (ESD) subsystem comprising a conventional safety logic solver, dedicated sensors, and actuators like manifold valves, safety valves, decoupling mechanism for pipeline connections \cite{handbookHydrogenVessels, europeanmaritimesafetyagencyHazardIdentificationGeneric2024, zhakaHydrogenFuelMaritime2024}. The ESD triggers automatically when any sensor or a voting group detects hydrogen above LFL, and can also be activated manually from the bunkering control room. This subsystem is entirely non-AI and satisfies the required functional safety integrity level on its own.

Segregated from the ESD, the AI-enabled LFL prediction system enhances operator situational awareness without compromising the independent safety function. It implements a two-stage pipeline: (1) an ensemble of MLPs estimates the unknown leakage rate, (2) an additive Gaussian process regressor forecasts the hydrogen dispersion field and its uncertainty over a short horizon. To reduce the risk of unnecessary barrier activations, an input guard checks sensor data integrity, e.g., missing values, format mismatches, flat-line signals. This input guard works based on rules that are manually programmed by a domain expert. Next, an AI monitoring component flags out-of-distribution inputs and unreliable internal behavior via explainable AI techniques \cite{flehmigPerspectivesReliabilityMonitoring2026}. The bunkering control room operator retains ultimate authority and the intrinsic safety of the system is given. The AI system outputs are advisory, and any safety barriers are triggered manually, while the ESD gets triggered automatically if the LFL is exceeded. Based on our interpretation of ISO/IEC TR 5469 \cite{iso/iectr5469:2024ArtificialIntelligenceFunctional2024}, the proposed architecture is intended to satisfy the criteria of usage level "D" where the intrinsic safety of the system is not disrupted by the AI system.

The GP forecast can be translated into a dynamic advisory display for the control room operators. Because the system is usage level "D", it is strictly advisory. It recommends, but never executes, any action. The predicted $H_2$ concentration over the short forecast horizon is mapped to three zones:
\begin{itemize}
    \item \textbf{Below LFL:} The operator receives a trend indication and continues normal monitoring.
    \item \textbf{Approaching LFL:} The system recommends preventive barriers, such as reducing the bunkering rate, water sprinklers, active ventilation. The operator decides whether to act.
    \item \textbf{Reaching LFL:} If the physical sensors subsequently exceed the LFL, the independent ESD system triggers automatically.
\end{itemize}
The AI system is intended to provide earlier awareness of hazardous concentration development before the ESD threshold is reached, thereby avoiding operational disruption while preserving intrinsic safety.

The proven-in-use approach requires evidence on conditions of use, failure-rate history, and impact analysis. In our framework, the operational envelope is defined by the training dataset and enforced at runtime by the input guard and AI monitoring components.  The low failure rates required for safety-related systems remain unachievable for data-driven models, training and validation metrics cannot replace operational evidence, but may serve as preliminary indicators prior to field deployment. Continuous reporting can be implemented and over time the operational evidence can support the claim that failure rates has not been exceeded. A conservative impact analysis must acknowledge that extrapolation beyond the tested leakage-rate range. For instance, at $\hat{m} > 1.0$\,g/s where data are sparse. It is unverified and should trigger manual supervision.

\subsection{Leakage rate estimation}\label{sec32}

\begin{figure}
    \centering
    \includegraphics[width=0.95\linewidth]{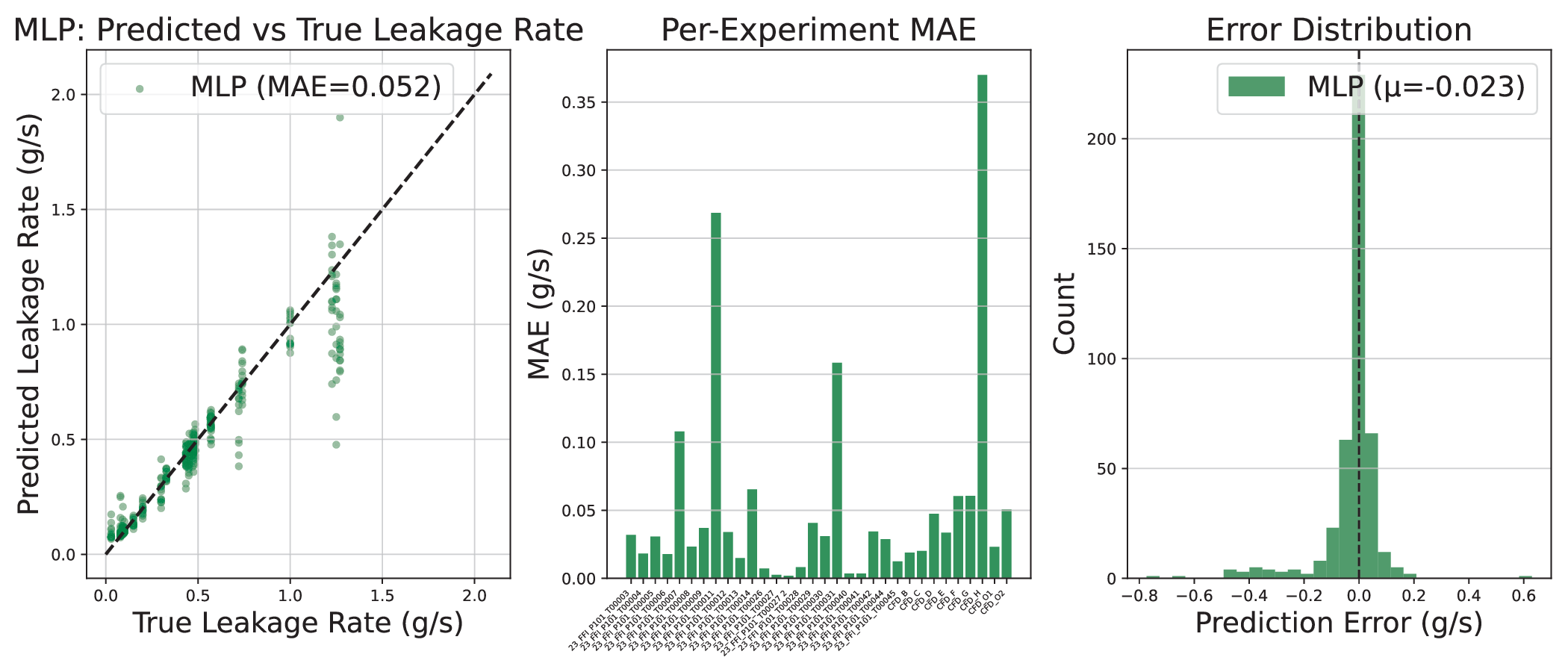}
    \caption{Performance of the Stage~1 MLP ensemble for leakage rate estimation. \textbf{Left:} Predicted versus true mass flow for all test-set samples (MAE = 0.052\,g/s); the dashed line marks perfect agreement.  \textbf{Middle:} Per-experiment MAE, showing reliable estimation for low-to-moderate leaks but elevated error for a few high-flow scenarios. \textbf{Right:} Distribution of prediction errors (true minus predicted), with mean $\mu = -0.023$\,g/s; the negative bias indicates a slight tendency toward under-prediction.}
    \label{fig:MassFlowEst}
\end{figure}
A MLP ensemble was developed to estimate the leakage rate $\hat{m}$ from early-time sensor observations. The input comprises 87 log-transformed sensor values, 29 sensors over three consecutive time steps, concatenated with 15 handcrafted spatiotemporal features. Each ensemble member is a 3-layer MLP with ReLU activations and dropout. To address the skewed leakage rate distribution, inverse-frequency sample weights were applied across leakage rate quantile bins. An ensemble of $N=7$ independently initialized networks was used. Predictions were averaged and the ensemble standard deviation served as an epistemic uncertainty estimate. As shown in Figure~\ref{fig:MassFlowEst}, the ensemble achieved a MAE of 0.052\,$g/s$, and a RMSE of 0.111\,$g/s$.

\subsection{Hydrogen dispersion forecast}\label{sec33}
\subsubsection{Full kernel screening with VNNGP}\label{sec331}
We first trained VNNGP models with the full additive kernel (Eq.~\ref{eq:general_additive}) to establish a performance baseline and identify which interaction terms dominate. The best VNNGP configuration considering MAE, RMSE and predictive log likelihood is $k=64$, LR $=0.008$, and a Gaussian likelihood. It achieved a validation MAE of $0.0201$, RMSE of $0.0277$ and predictive log likelihood of $-0.0166$ (Table~\ref{tab:scaleAdditive_twoModes_results}). The training was slow and the kernel structure was difficult to interpret.

Analysis of the learned output scales (Table~\ref{tab:kernel_terms_compact}) revealed a clear pattern. 18 kernels had near-zero scales with $\theta < 10^{-3}$, while the dominant contributions came from interactions involving both mass flow and sensor coordinates. This motivated the construction of a "light" additive kernel retaining only the 12 most dominant terms (Eq.~\ref{eq:light_additive_kernel}).

\subsubsection{Light kernel refinement with VNNGP and SVGP}\label{sec332}

With the light additive kernel, both VNNGP and SVGP were trained across different hyperparameters. Table~\ref{tab:light_Additive_twoModes_results} summarizes the VNNGP validation results. The light additive kernel improved training speed and predictive performance uniformly. For instance, VNNGP with $k=16$ and $\text{LR}=0.009$ achieved a validation MAE of $0.0136$ and RMSE of $0.0193$, outperforming the best full-kernel VNNGP model. SVGP models trained with $4000$--$6000$ inducing points further improved accuracy. The best SVGP configuration in terms of point estimation with $5000$ inducing points, RBF kernel, and Gaussian likelihood achieved a validation MAE of $0.0116$, RMSE of $0.0156$, and a predictive log-likelihood of $-0.2309$ (Table~\ref{tab:light_Additive_SVGP_val}). The best SVGP configuration in terms of predictive log-likelihood had $6000$ inducing points, a Beta likelihood, and an aggressive learning rate of $0.085$. It achieved a MAE of $0.0179$, RMSE of $0.0235$, and a predictive log-likelihood of $-2.6051$.

\subsubsection{Likelihood selection: Gaussian vs. Beta}\label{sec333}
\begin{figure}
    \centering
    \begin{subfigure}{\linewidth}
        \centering
        \includegraphics[width=0.95\linewidth]{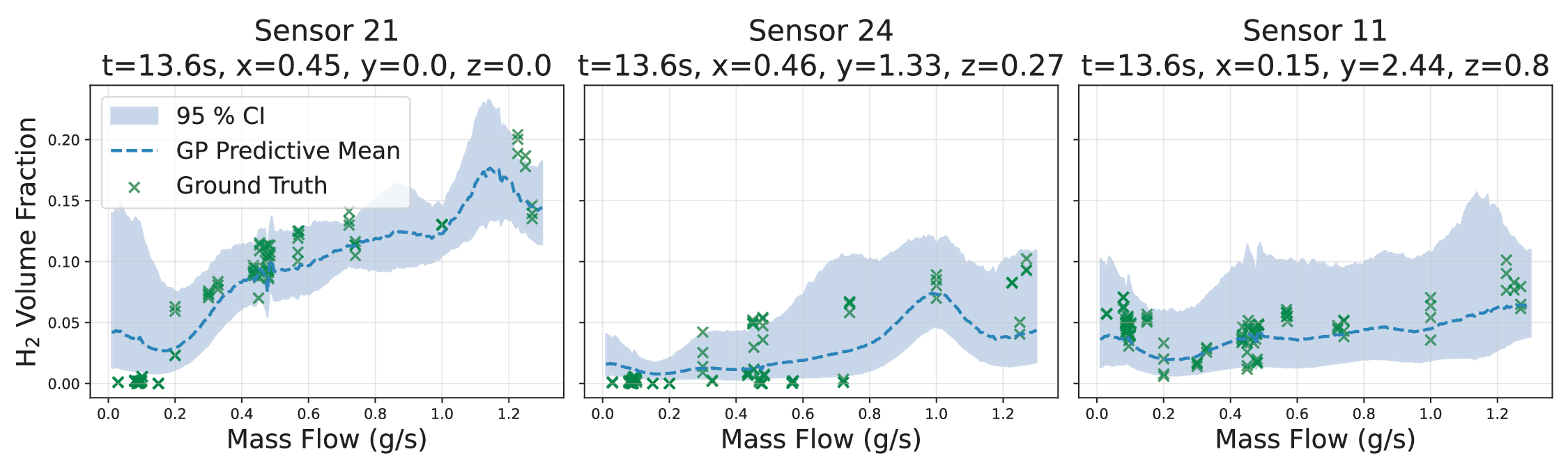}
        \caption{SVGP, Gaussian likelihood, $M=6000$}
        \label{fig:gaussian}
    \end{subfigure}
    
    \vspace{0.3cm}
    
    \begin{subfigure}{\linewidth}
        \centering
        \includegraphics[width=0.95\linewidth]{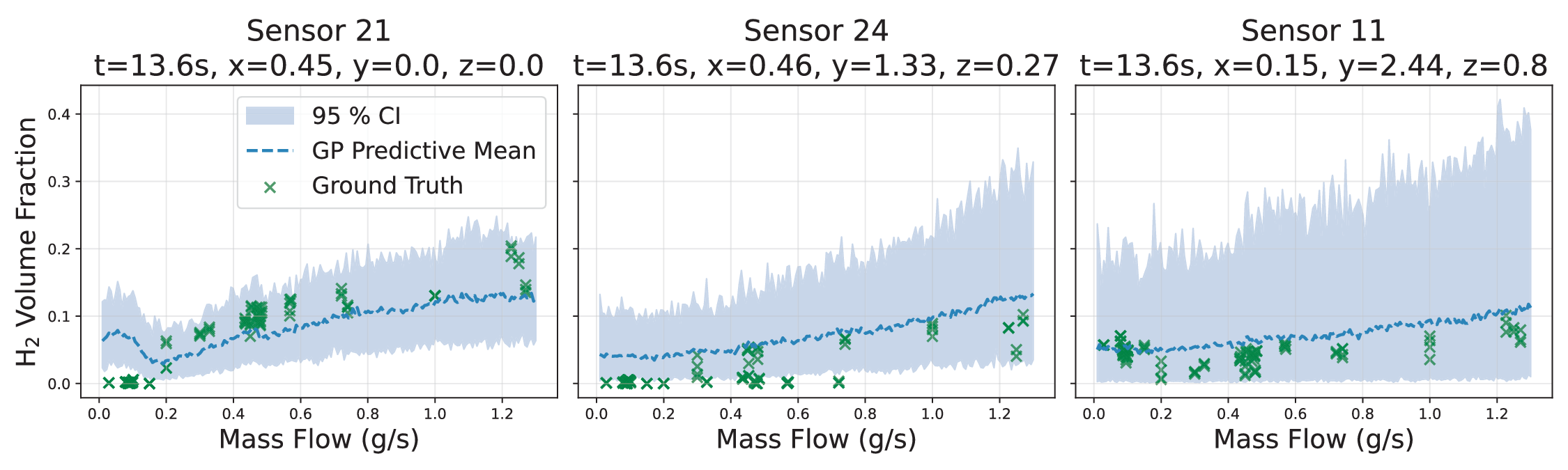}
        \caption{SVGP, Beta likelihood, $M=6000$}
        \label{fig:beta}
    \end{subfigure}
    \caption{Predictive mean (solid line) and 95\% confidence interval (shaded) for the two best SVGP models at three sensor positions, with ground truth (green crosses). \textbf{(a)}~Gaussian likelihood: accurate mean but overconfident uncertainty. \textbf{(b)}~Beta likelihood: wider, better-calibrated bands that capture the ground truth more reliably, especially at $\hat{m}\approx 1.0$\,g/s where training data are sparse.}
    \label{fig:likelihood_comparison}
\end{figure}

The target is a volume fraction bounded in $[0,1]$, we evaluated both Gaussian likelihood on log-transformed data and Beta likelihood on non-transformed data. Under VNNGP, the Beta likelihood performed poorly with the full additive kernel with a MAE of $0.0944$, and a RMSE of $0.1075$. It remained unsatisfactory with the light additive kernel with a MAE of $0.4206$, a RMSE of $0.4222$, and a predictive log-likelihood of $0.8641$. We attribute this to the sparsity of the VNNGP approximation, which struggles with the non-Gaussian observation model.

Under SVGP, however, the Beta likelihood performed well. With the light additive kernel and $4000$ inducing points, SVGP achieved a validation MAE of $0.0128$, RMSE of $0.0169$, and a predictive log likelihood of $-2.5602$. With $6000$ inducing points and an aggressive learning rate of $0.085$, the predictive likelihood performed even better with $-2.6051$. Models trained with Beta likelihood were substantially better calibrated than the Gaussian likelihood models, which exhibited over-confident predictions and high negative log-likelihood (NLL) (Table~\ref{tab:light_Additive_SVGP_val}). The NLL can be interpreted as measurement for calibration where a smaller value means better calibration. Figure~\ref{fig:likelihood_comparison} compares the two SVGP with different likelihoods at three representative sensors. The Gaussian model (a) achieves a slightly tighter mean fit but its 95\% bands are overconfident, frequently missing the ground truth at $\hat{m}$>0.8\,$g/s$, The Beta model (b) trades a marginal increase in MAE for better calibration, as reflected in its NLL of $-2.6051$ versus 12.6995.

\subsubsection{Final model and test-set evaluation}\label{sec334}
\begin{table}[htbp]
\centering
\setlength{\tabcolsep}{4pt}
\caption{Test-set performance of the best Stage-2 GP models with light additive kernel.}
\label{tab:gp_test_comparison}
\begin{tabular}{@{}llccc@{}}
\toprule
Method & Likelihood & MAE & RMSE & NLL \\
\midrule
VNNGP ($k=16$)  & Gaussian & 0.0248 & 0.0359 & 17.5418 \\
SVGP ($M=6000$) & Gaussian & 0.0207 & 0.0255 & 12.6712 \\
SVGP ($M=4000$) & Beta     & 0.0209 & 0.0267 & $-1.2255$ \\
SVGP ($M=6000$) & Beta     & \textbf{0.0176} & \textbf{0.0226} & $\mathbf{-1.5340}$ \\
\bottomrule
\end{tabular}
\end{table}
We tested the best VNNGP model, the best Gaussian SVGP model and the two best Beta SVGP models on the unseen test set, see Table~\ref{tab:gp_test_comparison}. Based on this, we selected the SVGP model with the light additive kernel, RBF base kernels, $6000$ inducing points, and Beta likelihood as the final Stage-2 regressor. On the held-out test set, this model achieved a MAE of $0.0176$, RMSE of $0.0226$, and NLL of $-1.5340$ (Table~\ref{tab:light_Additive_test_perf}). It outperformed all other tested models. The test-set MAE is clearly below a factor of $2$ of the sensor precision ($0.01$), indicating that the model approaches the fundamental measurement limit of the experimental setup.

The optimized kernel hyperparameters are listed in Table~\ref{tab:kernel_hyperparameters}. The dominant combinations confirm the physical intuition that spatial position, particularly $y$ and $z$, and leakage rate $\hat{m}$ are the primary drivers of concentration. Time has a comparatively weak effect after the initial transient, consistent with the quasi-steady-state behavior observed in Figure~\ref{fig:ConcentrationRampUp}.

\subsubsection{Calibration of final model}
\begin{figure}
    \centering
    \includegraphics[width=0.89\linewidth]{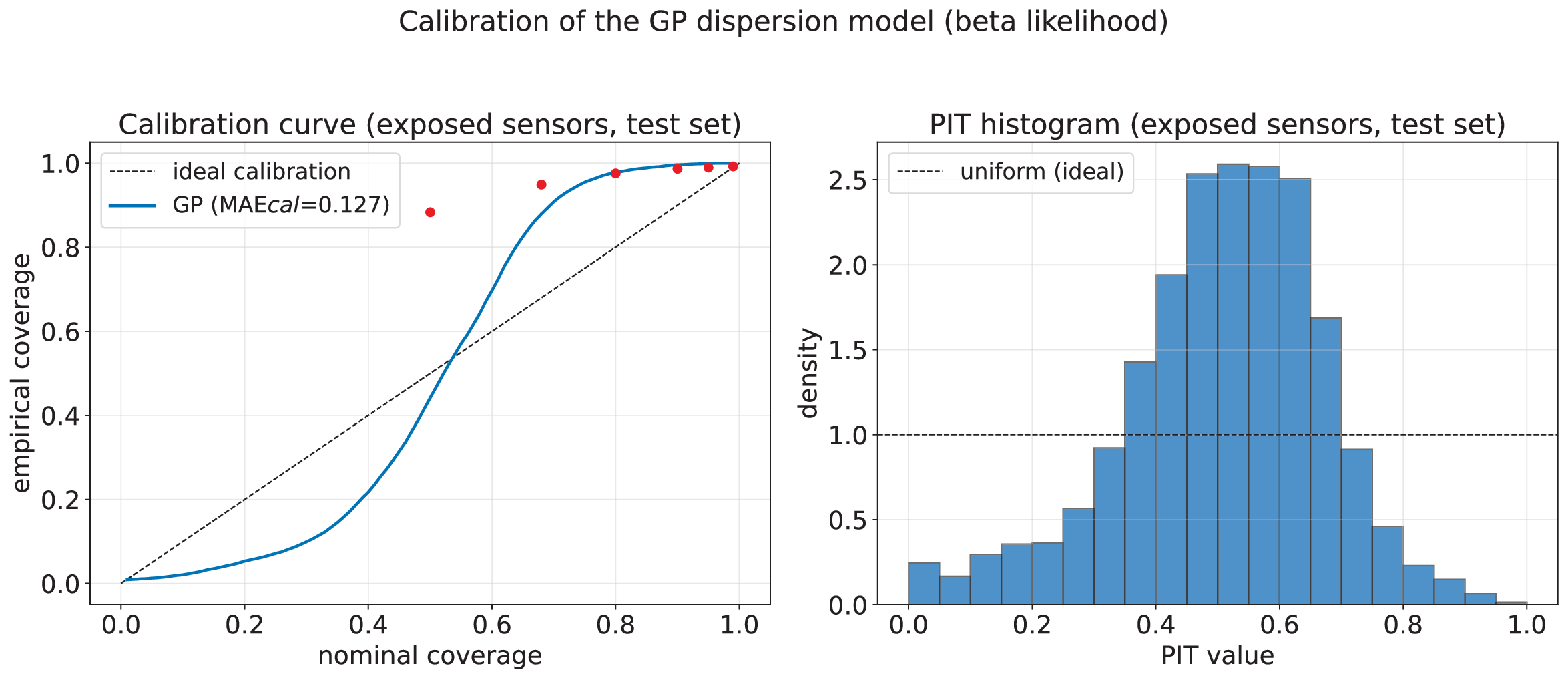}
    \caption{Calibration of the GP dispersion model (Beta likelihood) on the held-out test experiments (exposed sensor readings, n = 14,694; 1000 predictive Monte Carlo samples per point). (a) Empirical versus nominal coverage of the predictive quantiles. Red markers show PICP for the central 50–99\% intervals. The curve above the diagonal indicates conservative prediction intervals. (b) PIT histogram centered at 0.508 but narrower than uniform, confirming mild over-coverage.}
    \label{fig:gp_calibration}
\end{figure}
To assess the reliability of the predictive uncertainties, we evaluated the calibration of the final GP model on the four held-out test experiments. At each test point, 1000 Monte Carlo samples were drawn from the full predictive distribution, comprising both the posterior uncertainty of the latent GP and the observation noise of the Beta likelihood. Table~\ref{tab:coverage} reports the prediction interval coverage probability (PICP) for central prediction intervals at nominal levels between 50\% and 99\%, together with the corresponding mean prediction interval widths (MPIW). All statistics are computed on exposed sensor readings, i.e., the distribution the GP was trained on. The empirical coverage meets or exceeds the nominal coverage at every level. The 95\% interval contains 99.0\% of the observed concentrations and the 99\% interval 99.3\%. The probability integral transform (PIT) values are centered at 0.51, indicating negligible systematic bias, but their spread is narrower than that of a uniform distribution, and the calibration curve (Figure~\ref{fig:gp_calibration}) lies above the diagonal over most of the range while remaining nearly exact at the extreme quantiles. The predictive intervals are therefore conservative.

\begin{table}[htbp]
\centering
\small
\caption{Calibration results showing nominal coverage, empirical coverage, and MPIW.}
\label{tab:coverage}
\begin{tabular}{@{}ccc@{}}
\toprule
Nominal coverage & Empirical coverage & MPIW [conc.] \\
\midrule
0.50 & 0.883 & 0.044 \\
0.68 & 0.949 & 0.065 \\
0.80 & 0.974 & 0.084 \\
0.90 & 0.987 & 0.108 \\
0.95 & 0.990 & 0.128 \\
0.99 & 0.993 & 0.166 \\
\bottomrule
\end{tabular}
\end{table}

\subsubsection{Sensor threshold sensitivity}

The threshold of $0.009$ for exposed sensors was selected heuristically based on the sensor precision of $\pm0.01$ without  detailed analysis of its suitability. To assess the sensitivity of the GP model to this choice, we retrained the final SVGP configuration with Beta likelihood and 6000 inducing points using alternative thresholds of 0.0095 and 0.0085. Table~\ref{tab:threshold_datasets} summarizes how these adjustments affect the exposed-only dataset. The number of exposed samples changes only marginally, and this variation propagates to the model predictions. Neither alternative threshold yields significant deviations from the default configuration in terms of MAE, RMSE, or predictive log-likelihood. For the 0.0085 threshold, the model achieves a MAE of 0.0198, an RMSE of 0.0270, and a predictive log-likelihood of $-2.436$. For 0.0095 threshold, the corresponding values are 0.0186, 0.0254, and $-2.532$.

\begin{table}[htbp]
\centering
\small
\setlength{\tabcolsep}{4pt}
\caption{Sample counts for datasets generated with different exposure thresholds. The exposed share is 64.1\,\%, 63.8\,\%, and 63.6\,\% for thresholds 0.0085, 0.0090, and 0.0095, respectively.}
\label{tab:threshold_datasets}
\begin{tabular}{@{}l r r r r r@{}}
 \toprule
 Threshold & Total & Exposed & Train. & Val. & Test \\
 \midrule
 0.0085 & 182\,181 & 116\,768 & 78\,444 & 23\,589 & 14\,735 \\
 \textbf{0.0090} & \textbf{182\,181} & \textbf{116\,306} & \textbf{78\,133} & \textbf{23\,479} & \textbf{14\,694} \\
 0.0095 & 182\,181 & 115\,868 & 77\,835 & 23\,388 & 14\,645 \\
 \bottomrule
\end{tabular}
\end{table}

\subsection{End-to-end framework evaluation}\label{sec34}
During runtime, the MLP ensemble and GP model operate sequentially as outlined in Algorithm~\ref{alg:runtime}, meaning errors in the MLP's leakage rate estimation propagate to the hydrogen forecasting stage. To quantify this effect, we compared our oracle model where the GP receives ground-truth mass flow from the held-out test sets (Section~\ref{sec334}) against the same data using the MLP ensemble's estimated leakage rate as input. As shown in Table~\ref{tab:twoStageperformance}, this substitution increases MAE from $0.0176 \rightarrow 0.0249$ with corresponding increases in RMSE and NLL. 

\begin{table}[htbp]
\centering
\small
\setlength{\tabcolsep}{4pt}
\caption{Performance comparison of two-stage approach versus oracle.}
\label{tab:twoStageperformance}
\begin{tabular}{@{}lccc@{}}
    \toprule
    Configuration & MAE & RMSE & NLL \\
    \midrule
    Two-stage (MLP est.\ $\hat{m}$) & 0.0249 & 0.0318 & $-1.421$ \\
    Oracle (true $m$) & 0.0176 & 0.0236 & $-1.534$ \\
    \bottomrule
\end{tabular}
\end{table}

\section{Discussion}\label{sec4}

\subsection{System design and compliance with policies}\label{sec41}
The proposed system design follows ISO/IEC TR 5469 in classifying AI system as usage level "D" and in adopting its architectural pattern. This classification makes established risk reduction measures from functional safety standards like ISO 61508 applicable, thereby lowering the barrier for incorporation of AI systems into a safety-critical system. The proposed design must be regarded as an initial baseline rather than a final implementation. It marks the starting point for an iterative risk assessment process through which the design will be refined.

As an initial design, it is still meaningful because it embeds core safety and reliability principles. The independent ESD remains the ultimate safety layer, the AI system's reliability is intended to be increased by architectural elements such as the input guard, and the AI monitoring. These components can be viewed as minimum requirements for any AI system that augments a safety-critical process.

For the overall system, additional conventional safeguards can be implemented. Conservative alarm thresholds in high-uncertainty regions and rejection of predictions that fall outside the system's competence envelope, as flagged by the monitoring components.

In addition to adequate system design that promotes safe and reliable operation, human operators must also be prepared to interpret outputs from both the AI system and the independent monitoring components, i.e., input guard and AI monitoring. This preparation is essential to realize the full benefit of the AI-enabled framework without compromising the safety integrity or operational effectiveness of the independent ESD system.

This paper is not the first to draw on the “proven-in-use” approach from software safety standards. The ISO/IEC TR 5469 \cite{iso/iectr5469:2024ArtificialIntelligenceFunctional2024} already mentions this approach but does not further elaborate. This works proposes a perspective on how proven-in-use can be translated to AI systems in safety-related systems where the training dataset represents the operational envelope, and AI monitoring and input guard ensure that the system stays within this envelope. Whereas the translation is sound, the approach needs to be evaluated in more depth.

\subsection{Uncertainty-aware AI systems and human operators}
Implementing uncertainty-aware AI systems does not come for free. Displaying not just point estimates but also their associated uncertainty constitutes a powerful tool when correctly implemented and, equally or even more importantly, correctly interpreted. Thus, operators should ideally participate in the development process of such AI systems from the start to develop an intuitive understanding of their functionality. Where this proves impractical, operators must receive training to interpret uncertainty estimates correctly, or an additional layer must be implemented that flags point estimates falling within high-uncertainty regions as potentially unreliable.

Operators should interpret point estimates with high uncertainty cautiously and act conservatively. They should rather trigger a safety barrier than refraining from action. In terms of safety integrity, no additional risk is introduced, and the independent ESD system remains the ultimate safety layer.

\subsection{Tank connection space geometry}
Even with only data from a sparse sensor network, the two-stage framework demonstrates that it can learn hydrogen dispersion physics from real-world measurements and CFD simulations and deliver calibrated uncertainty in real time. These are important capabilities required for a bunkering safety-related system, independent of the exact geometry.

The open-ended rectangular channel used in this work possesses properties transferable to a ship's tank connection space. Leakages typically arise at the connection between the pipeline and the receiving ship. Thus, the assumption of a single, localized leakage source as employed in this work remains realistic. Furthermore, the open-ended channel reflects the natural ventilation common in such spaces. However, while a tank connection space could theoretically be designed similarly to this setup, it would more realistically incorporate active ventilation functioning as a safety barrier and contain obstructions that influence flow patterns. Consequently, the model trained in this work cannot be directly transferred to such geometries.

The aim of this work was not to build a model that generalizes to all geometries, but rather one that is easily re-trainable on new configurations. The MLP feature extraction depends on sensor topology rather than specific geometry, meaning it can be readily re-trained on new geometries. The same applies to the additive GP kernel. A complete subset kernel as in Eq.~\ref{eq:general_additive} can initially be trained, after which unimportant kernels can be discarded. Future work should validate the framework on different geometries to test this transferability. One specific challenge is the introduction of active safety barriers, as these represent new runtime inputs that would require a model extension to address this additional dimension.

\subsection{Dataset size, distribution skew, and scalability trade-offs}
The dataset comprises only 32 unique scenarios. This is modest by conventional machine-learning standards but it is representative of the operational reality in safety-critical process industries. Hazardous events are rare by design, pre-operational systems are not yet deployed at scale, and large-scale destructive experiments are expensive. Simulation data are approximations, and physical phenomena such as hydrogen dispersion remain challenging to model completely. This work therefore addresses a practical problem with extracting maximum value from small, expensive experimental data. Consequently, the framework was deliberately built to avoid large models. The additive GP structure was selected precisely because GPs offer robust generalization with small-to-medium datasets and provide predictive uncertainty, which is essential when training data cannot cover the full operational envelope. The MLP ensemble is similarly compact, and the feature engineering reduces dimensionality to match the sample size.

The empirical leakage rate distribution is heavy-tailed, with dominant peaks near 0.1 $g/s$ and 0.4--0.5 $g/s$ and sparse coverage above 0.8 $g/s$, see Figure~\ref{fig:massFlow}. This directly degrades estimation accuracy at the extremes, see Figure~\ref{eq:mlp_metrics} and widens GP uncertainty bands where data are scarce, see Figure~\ref{fig:likelihood_comparison}. The system design explicitly compensates for it with higher uncertainties in these input spaces. Additionally, AI monitoring and input guard flag inputs that are outside the learned distribution or violate the data integrity. Future operational deployment should employ active learning or targeted CFD augmentation to populate the 0.8--1.3 $g/s$ regime. Consequently, the present results should be interpreted as a proof-of-concept demonstration of the proposed framework rather than evidence of deployment readiness in operational maritime bunkering systems.

Expanding the dataset through additional CFD simulations, experimental data, or higher-density sensors with increased measurement frequency would initially improve performance. GPs can be scaled to accommodate larger datasets with more powerful GPUs, but computational limits emerge with very large datasets. This presents a trade-off between either restrict the dataset to preserve the GP's interpretability and uncertainty quantification, or transition to models designed for large-scale data, such as deep learning approaches.

\subsection{Model selection}\label{sec43}
The first-stage estimator was chosen for computational speed at inference, ease of implementation, and sufficient accuracy. The ensemble structure provides two additional benefits. A slight performance gain over single networks and a heuristic uncertainty estimate derived from the variance across members, which is practically useful for flagging unreliable leakage rate predictions to operators.

\begin{figure}
    \centering
    \includegraphics[width=0.9\linewidth]{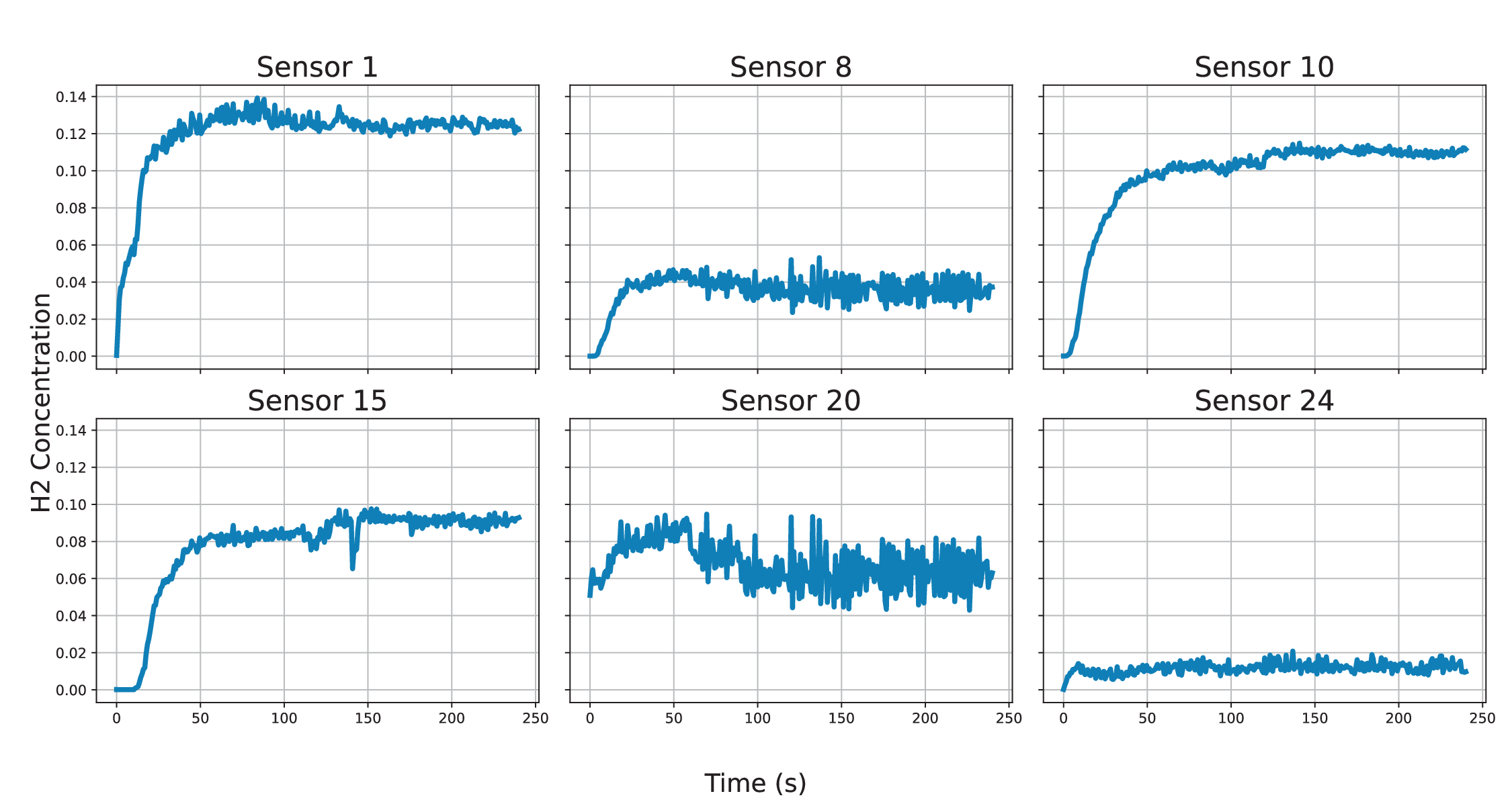}
    \caption{Hydrogen concentration measured at six representative sensors (1, 8, 10, 15, 20, 24) over all tests. All sensors exhibit a rapid transient rise during the first approximately 50--100~s, followed by a quasi-steady-state plateau. The post-transient behavior is largely independent of time, indicating that temporal evolution is primarily relevant during the initial dispersion phase. For the majority of the leakage duration, sensor position and mass flow rate dominate the concentration mapping.}
    \label{fig:ConcentrationRampUp}
\end{figure}

The second-stage model was selected because the limited dataset makes deep learning approaches prone to overfitting, whereas GPs offer robust generalization with small-to-medium-sized datasets. More importantly, GPs provide predictive uncertainty and an interpretable kernel structure, properties that are especially valuable in safety-critical applications where operators must understand and trust model outputs. The final additive kernel comprises 12 interaction terms with 12 output scales and 33 lengthscale parameters (see Eq.~\ref{eq:light_additive_kernel}). Selecting the 12 most dominant terms was an intuitive choice. The first 11 had almost identical output scales, and the $12^{th}$ was retained because it was the only term containing the time dimension. This structure may be too complicated to interpret at runtime, but offline inspection before deployment is feasible. Consequently, model maintenance is easier because the parameter set is compact and each term is individually understandable.

Intuitively, we expected time to be more important for the prediction. We explain this by the near-steady-state concentration reached after an initial ramp-up (Figure~\ref{fig:ConcentrationRampUp}). Shortly after leakage onset, time is an important feature together with mass flow and sensor position. After approximately 50~s, however, the effect of time on the $H_2$ concentration has largely vanished and the concentration becomes almost constant.

From a safety perspective, training of data-driven models is probabilistic and thus cannot satisfy the extremely low failure rates typically required of safety functions. The integrated uncertainty quantification and interpretable kernel structure of GPs, see Figure~\ref{fig:likelihood_comparison}(b), do not close this gap, but they support the detection of unreliable behavior and thereby reduce risk. Interpretability also benefits long-term maintenance where flaws in model behavior are easier to diagnose.

With substantially more data, deeper architectures could be considered. DeepONet \cite{luDeepONetLearningNonlinear2021}, for example, excels at learning the solution operators of partial differential equations and would be well suited to the hydrogen dispersion physics. For leakage-rate estimation, a graph neural network \cite{sanchez-lengelingGentleIntroductionGraph2021} could exploit the irregular sensor network topology. Such models might improve predictive accuracy and practical reliability, but they do not automatically enhance system safety. Increased complexity often reduces interpretability, which can degrade operator trust and situational awareness, effects that contradict the very purpose of their implementation.

\subsection{MLP ensemble for leakage estimation}\label{sec441}
\begin{figure}
    \centering
    \includegraphics[width=0.9\linewidth]{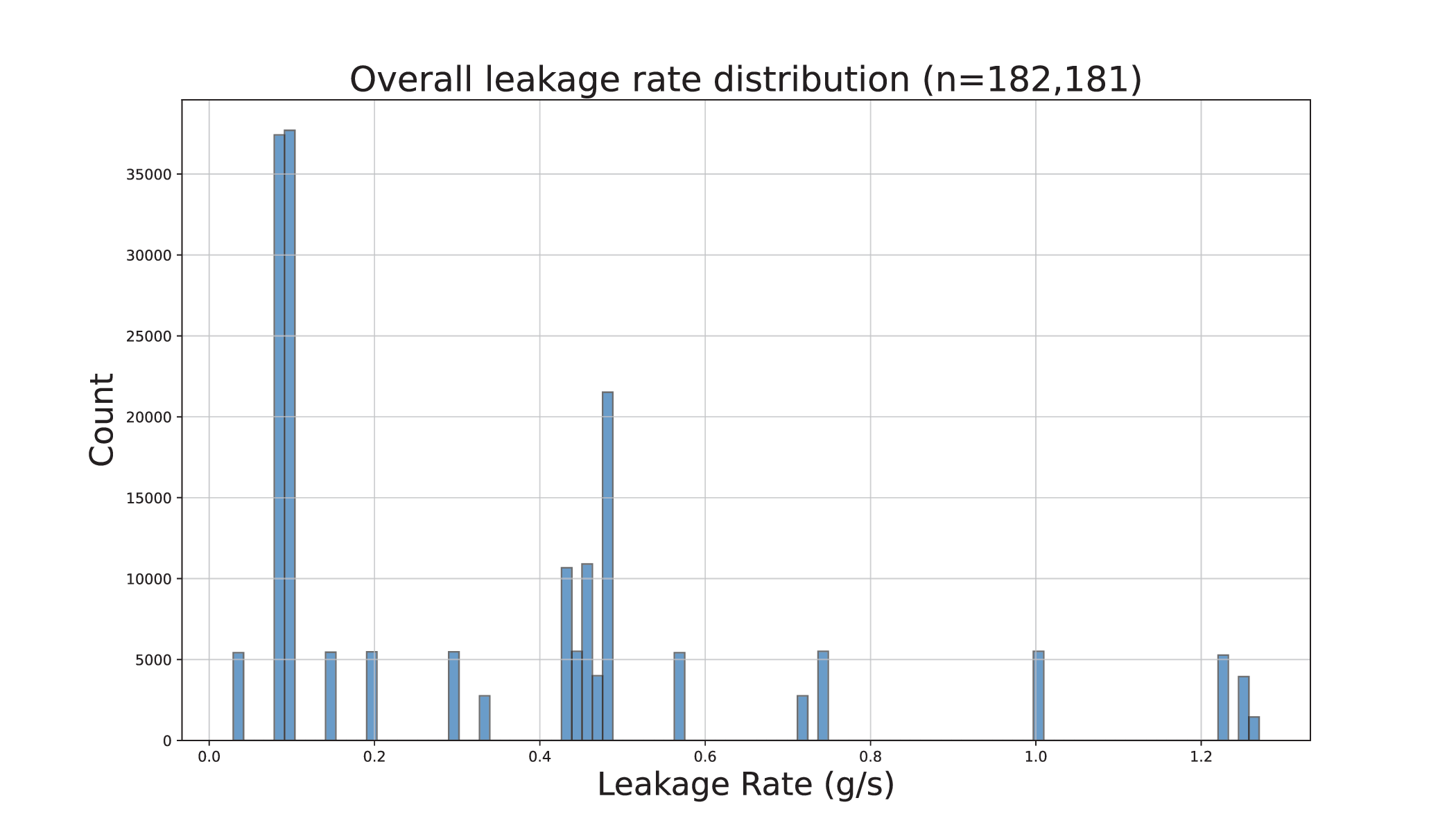}
    \caption{Overall distribution of leakage rates in the trimmed dataset. The empirical distribution is heavy-tailed, with dominant peaks near 0.1\,g/s and in the 0.4--0.5\,g/s band. Data become scarce above 0.8\,g/s and at the lowest rates, which limits model accuracy at the extremes.}
    \label{fig:massFlow}
\end{figure}
The prediction accuracy was strongly dependent on the leakage rate distribution that is shown in Figure~\ref{fig:massFlow}. For low-to-moderate leaks $\hat{m} < 0.7\,g/s$, where the majority of data resides, the MAE remains low and the model reliably captured the leak magnitude. In contrast, high-flow experiments $\hat{m} > 1.0\,g/s$ exhibited severe under-prediction. The MLP struggles with extreme leaks because of data scarcity in this region. The bottleneck for the performance is less the model architecture but rather the training data coverage.

\subsection{GP regression for hydrogen dispersion}\label{sec442}
The results in Section~\ref{sec33} confirm that the choice of approximation method and likelihood is tightly coupled to the safety-relevant properties of the model.

VNNGP served as an effective screening tool. Its relatively fast training allowed exploration of kernel hyperparameters and led to the identification of the 12 dominant interaction terms. However, VNNGP performance did not improve with larger $k$. Indeed, it degraded or plateaued (Section~\ref{sec332}). We attribute this to data sparsity in the five-dimensional input space, where a large neighborhood radius includes points that are not close in feature space. Moreover, VNNGP is designed for spatiotemporal problems where global approximations fail. Our relatively small spatial domain is adequately described by global inducing points, which explains why SVGP ultimately produced superior accuracy.

The likelihood comparison (Section~\ref{sec333}) reveals that domain knowledge and the correct choice of likelihood is important for the performance. Gaussian likelihood models achieved good point estimates on the test set, but their NLL was orders of magnitude worse than the Beta likelihood models (Table~\ref{tab:gp_test_comparison}). This indicates over-confidence. The Gaussian models produce narrow uncertainty bands that frequently exclude the ground truth, especially at high mass flow where data are sparse (Figure~\ref{fig:likelihood_comparison}a). In contrast, the Beta likelihood models respect the $[0,1]$ bound of the physical variable, yielding well-calibrated uncertainty that widens appropriately in extrapolation regimes (Figure~\ref{fig:likelihood_comparison}b). For runtime operation, this calibration is important. An operator who sees a tight but wrong confidence interval will trust the model when it is unreliable, whereas a slightly wider but honest interval triggers justified caution.

The final SVGP model with Beta likelihood approaches the sensor precision bound (MAE $0.0176$ vs.\ sensor accuracy $0.01$). This suggests that further point-estimation gains require either improved sensors or additional data, not merely model complexity. The kernel hyperparameters (Table~\ref{tab:kernel_hyperparameters}) corroborate the physical intuition that spatial position and leakage rate dominate concentration after the initial transient, whereas time has a comparatively weak effect, consistent with the quasi-steady state observed in Figure~\ref{fig:ConcentrationRampUp}. This interpretable structure supports maintenance and proven-in-use documentation.

\subsection{Calibration}
Computing uncertainties alongside point estimates can inform operator decision-making, but only when these uncertainties are properly calibrated. Otherwise, they risk misleading operators and undermining trust. In this work, the model exhibits a tendency toward over-estimation with producing prediction intervals with empirical coverage exceeding nominal targets (Table~\ref{tab:coverage}). This mild over-coverage is acceptable for safety-critical dispersion forecasting, where under-confident predictions are preferable to over-confident ones. Interval sharpness remains high relative to the observed concentration range, ensuring the uncertainty estimates remain informative rather than excessively broad.

This conservative bias likely stems from two sources. First, the Beta likelihood's learned observation noise may be inflated by high-noise regions in the training data (e.g., near the source during early transients). Second, variational GP approximations employing many inducing points can overestimate posterior variance, further widening uncertainty intervals.

Nevertheless, the model remains unbiased, and its intervals never systematically exclude the truth more frequently than the nominal coverage promises. For hydrogen safety applications, a model that acknowledges uncertainty while achieving coverage exceeding its stated confidence is far more defensible than one that is over-confident and subsequently wrong.
   
\subsection{Two-stage framework and its hazard underestimation}
A critical safety concern is the propagation of Stage~1 errors into Stage~2 especially in high-leakage-rate scenarios. The GP regressor conditions its forecast on the estimated leakage rate $\hat{m}$, any MLP under-prediction at high leakage rates, as seen in Figure~\ref{eq:mlp_metrics}, directly biases the GP toward lower predicted concentrations. The system would underestimate hazards when the leak is most severe. Results from the held-out test set (Section~\ref{sec34}) indicate that this error propagation increases MAE by approximately 1\% of hydrogen concentration. We consider this as acceptable, though improving Stage~1 reliability remains essential for ensuring Stage~2 robustness.

The system design mitigates this propagation risk through multiple safeguards. First, the input guard and AI monitoring component flag data integrity violations, out-of-distribution samples, and deviations from normal system behavior. Second, the MLP ensemble variance serves as a heuristic reliability flag where high uncertainty triggers another flag. Finally, the independent ESD system ensures that an automatic shutdown still occurs if the true concentration reaches LFL. Thus, the propagation risk is contained by the designed segregation, but it underscores why the AI system must remain advisory only.

\section{Conclusion}\label{sec5}
This work presented a two-stage, uncertainty-aware framework fore real-time prediction of hydrogen dispersion during bunkering in confined spaces. The first stage employs an MLP ensemble to estimate leakage rates from features extracted during the initial ten seconds after leak onset, whereas the second stage applies an additive Gaussian process to forecast the spatiotemporal concentration field with calibrated uncertainty. Predictions are computed in seconds rather than the hours required by CFD. By augmenting current sensor readings with forecasts of future hydrogen concentrations, the framework aims to enhance operator situational awareness and promotes a shift from reactive response to proactive management of safety barriers. The GP's uncertainty estimates allow operators to distinguish between trustworthy predictions from extrapolations beyond the training data, addressing a critical requirement for safety-critical decision support.

Methodologically, the adjusted additive kernel strategy, where a broad kernel search is followed by a refinement of the most dominant components, provides a practical heuristic for GP model selection and results in an interpretable final structure. The framework is embedded within a system design compliant with ISO/IEC TR 5469, featuring segregation from independent emergency shutdown system, input guard, and AI monitoring component.

Validation on experimental and CFD data demonstrates that the MLP ensemble estimates leakage rates with a mean absolute error of  $0.052$~g/s, whereas the GP achieves a mean absolute error of $\approx 0.0176$ in volume fraction, approaching the sensor precision bound of $\pm0.01$, with a calibrated NLL of $-1.5340$. Although accuracy degrades at extreme leakage rates due to data scarcity, these results confirm the framework's capability to augment sensor readings with interpretable, uncertainty-quantified predictions that support situational awareness, and thus decision making.

Several limitations suggest directions for future work. The current validation reflects a single, idealized geometry with skewed leakage rate distribution. Transfer to realistic ship geometries and diverse leak scenarios requires further experimental validation and dataset expansion, potentially via active learning targeted at under-represented regions. A quantitative comparison with reduced-order CFD implementations would also be valuable for benchmarking computational efficiency and forecasting accuracy. However, such models were not accessible within the scope of this work and remain an interesting direction for future investigation. The input guard and AI monitoring component require concrete implementation and deeper investigation. In addition, the adaption of the "proven-in-use" concept would need to be tested through extended operational validation to see whether safety assurance requirements can be satisfied. Finally, improving Stage~1 reliability is essential for deployment. Nevertheless, this framework provides a reproducible foundation for integrating interpretable, uncertainty-aware AI into safety-related systems for hydrogen bunkering.

\section*{Declaration of competing interest}
The authors declare that they have no known competing financial interests or personal relationships that could have appeared to influence the work reported in this paper.

\section*{Acknowledgment}
This work was carried out as a part of SUBPRO-Zero, a Research Centre at the Norwegian University of Science and Technology (project code 100589 and CristinID 2720670). The authors gratefully acknowledge the project support from SUBPRO-Zero, which is financed by  major industry partners and NTNU. We also thank Hannibal Fossum from the Norwegian Defence Research Establishment for helpful insights on the experiments and providing additional CFD data.

\section*{Data availability}
The AI system will be made available through a GitHub (\url{https://github.com/nicmig/H2-dispersion-model}) repository when the article is published.

\section*{Generative AI disclosure statement}
During the preparation of this manuscript, the authors used Kimi (K3), Microsoft Copilot, and ChatGPT solely for language editing and sentence refinement. The authors reviewed and edited all AI-assisted outputs and take full responsibility for the accuracy, integrity, and final content of the manuscript. No AI tool was used to generate original data, perform independent scientific interpretation, or replace author judgment.

\clearpage
\appendix
\section{Kernel definition}

\begin{subequations}
\begin{align}
k_{\text{add}}(\mathbf{x},\mathbf{x}') &= 
+ \theta_{1,2}k_{12}
+ \theta_{1,3}k_{13}
+ \theta_{1,4}k_{14}\nonumber\\[2pt]
&\quad 
+ \theta_{2,3}k_{23}
+ \theta_{2,4}k_{24}
+ \theta_{3,4}k_{34}\nonumber\\[2pt]
&\quad 
+ \theta_{1,2,3}k_{123}
+ \theta_{1,2,4}k_{124}
+ \theta_{1,3,4}k_{134}\nonumber\\[2pt]
&\quad 
+ \theta_{2,3,4}k_{234}
+ \theta_{1,2,3,4}k_{1234}\nonumber\\[2pt]
&\quad 
+ \theta_{0,1,2,3,4}k_{01234}.
\label{eq:light_additive_kernel}
\end{align}
\end{subequations}
Equation~\eqref{eq:light_additive_kernel} shows the final light additive kernel decomposition.

\vspace{\baselineskip}

\begin{center}
\footnotesize
\setlength{\tabcolsep}{3pt}
\captionof{table}{Optimized hyperparameters of the additive kernel terms (SVGP, Beta likelihood, $M=6000$). Lengthscales $(l_0,l_1,l_2,l_3,l_4)$ correspond to $(t,\hat{m},x,y,z)$. Empty cells denote inactive dimensions.}
\label{tab:kernel_hyperparameters}
\begin{tabular}{@{}ccccccccc@{}}
\toprule
Term & Order & $l_0$ & $l_1$ & $l_2$ & $l_3$ & $l_4$ & $\theta$ \\
\midrule
0  & 2 & --- & 4.02 & 4.24 & --- & --- & 0.453 \\
1  & 2 & --- & 3.65 & --- & 1.08 & --- & 0.135 \\
2  & 2 & --- & 3.72 & --- & --- & 2.94 & 0.605 \\
3  & 2 & --- & --- & 4.68 & 3.29 & --- & 0.614 \\
4  & 2 & --- & --- & 5.48 & --- & 2.19 & 0.636 \\
5  & 2 & --- & --- & --- & 1.92 & 1.84 & 0.218 \\
\midrule
6  & 3 & --- & 4.65 & 4.21 & 3.51 & --- & 0.417 \\
7  & 3 & --- & 0.06 & 12.65 & --- & 14.12 & 0.003 \\
8  & 3 & --- & 0.20 & --- & 0.48 & 0.70 & 0.036 \\
9  & 3 & --- & --- & 4.26 & 2.69 & 1.83 & 0.294 \\
\midrule
10 & 4 & --- & 2.25 & 2.49 & 1.07 & 0.91 & 0.140 \\
\midrule
11 & 5 & 0.43 & 3.80 & 33.49 & 2.31 & 3.21 & 0.114 \\
\bottomrule
\end{tabular}
\end{center}

\newpage

\section{Tables and figures}

\begin{center}
\small
\setlength{\tabcolsep}{4pt}
\captionof{table}{Additive kernel combinations and learned output scales $\theta$ for the five-dimensional input $(t, m, x, y, z) \equiv (0,1,2,3,4)$. Highlighted cells indicate the 12 dominant terms.}
\label{tab:kernel_terms_compact}
\begin{tabular}{@{}clr@{}}
\toprule
\textbf{Ord.} & \textit{Combination} & \textit{$\theta$} \\
\midrule
1 & $(0,)$       & 8.4542e-05 \\
1 & $(1,)$       & 3.2086e-03 \\
1 & $(2,)$       & 2.2089e-03 \\
1 & $(3,)$       & 1.6013e-03 \\
1 & $(4,)$       & 2.0619e-03 \\
2 & $(0, 1)$     & 2.9648e-04 \\
2 & $(0, 2)$     & 2.1476e-04 \\
2 & $(0, 3)$     & 2.6015e-04 \\
2 & $(0, 4)$     & 3.6644e-04 \\
2 & \cellcolor{blue!12}$(1, 2)$     & \cellcolor{blue!12}6.1894e+01 \\
2 & \cellcolor{blue!12}$(1, 3)$     & \cellcolor{blue!12}6.1559e+01 \\
2 & \cellcolor{blue!12}$(1, 4)$     & \cellcolor{blue!12}6.1912e+01 \\
2 & \cellcolor{blue!12}$(2, 3)$     & \cellcolor{blue!12}6.5205e+01 \\
2 & \cellcolor{blue!12}$(2, 4)$     & \cellcolor{blue!12}6.5305e+01 \\
2 & \cellcolor{blue!12}$(3, 4)$     & \cellcolor{blue!12}6.5262e+01 \\
3 & $(0, 1, 2)$    & 6.4878e-04 \\
3 & $(0, 1, 3)$    & 1.9283e-03 \\
3 & $(0, 1, 4)$    & 9.6029e-04 \\
3 & $(0, 2, 3)$    & 4.7215e-04 \\
3 & $(0, 2, 4)$    & 8.7390e-04 \\
3 & $(0, 3, 4)$    & 1.2251e-03 \\
3 & \cellcolor{blue!12}$(1, 2, 3)$    & \cellcolor{blue!12}6.2965e+01 \\
3 & \cellcolor{blue!12}$(1, 2, 4)$    & \cellcolor{blue!12}6.3147e+01 \\
3 & \cellcolor{blue!12}$(1, 3, 4)$    & \cellcolor{blue!12}6.2925e+01 \\
3 & \cellcolor{blue!12}$(2, 3, 4)$    & \cellcolor{blue!12}6.5294e+01 \\
4 & $(0, 1, 2, 3)$ & 4.1389e-03 \\
4 & $(0, 1, 2, 4)$ & 2.5979e-03 \\
4 & $(0, 1, 3, 4)$ & 1.8597e-02 \\
4 & $(0, 2, 3, 4)$ & 2.2726e-03 \\
4 & \cellcolor{blue!12}$(1, 2, 3, 4)$ & \cellcolor{blue!12}6.3641e+01 \\
5 & \cellcolor{blue!12}$(0, 1, 2, 3, 4)$ & \cellcolor{blue!12}8.2382e-02 \\
\bottomrule
\end{tabular}
\end{center}

\vspace{0.5\baselineskip}

\begin{center}
\includegraphics[width=0.7\linewidth]{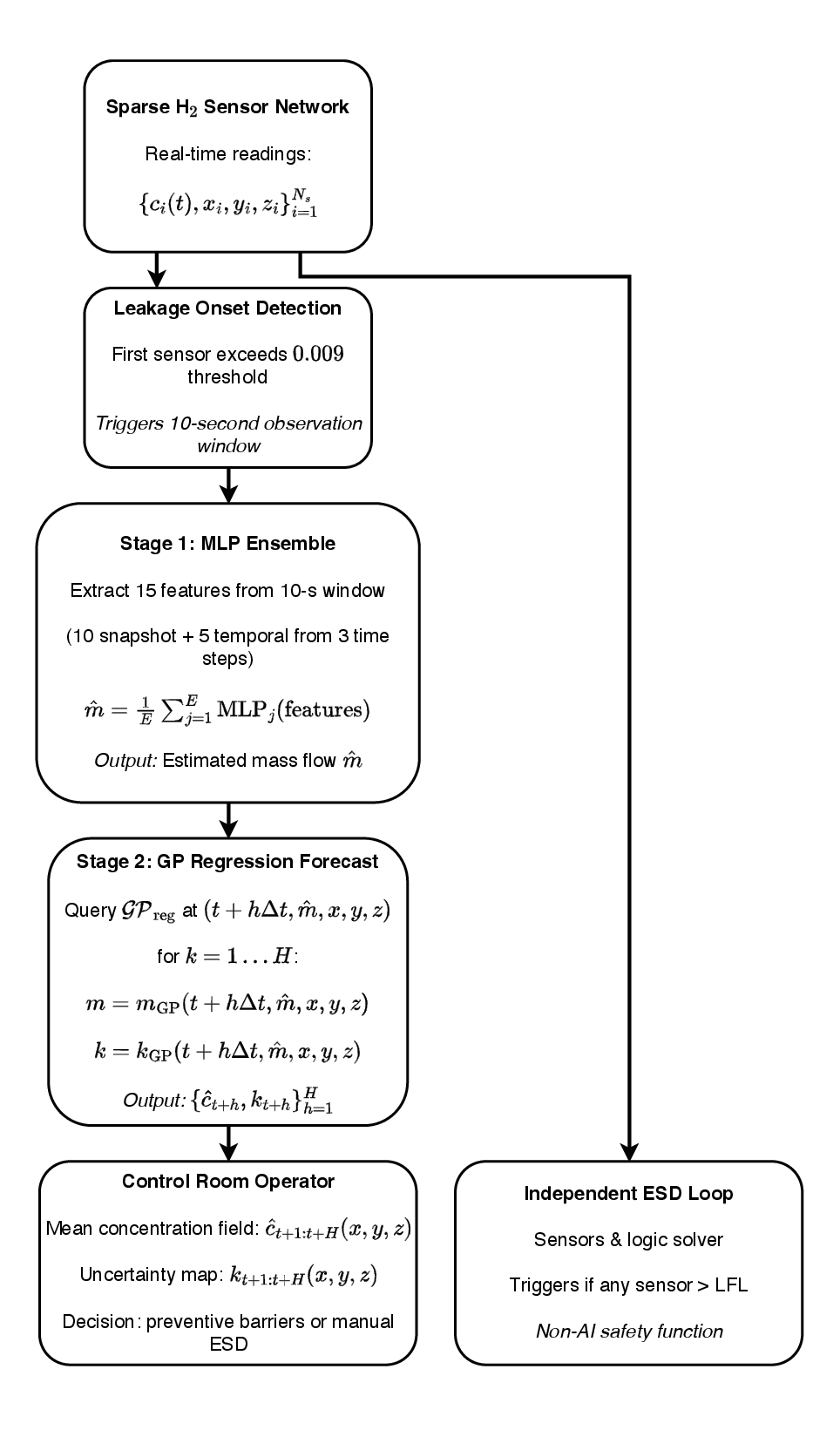}
\captionof{figure}{Simple flowchart showing the runtime framework.}
\label{fig:GP_EKF_flow}
\end{center}

\vspace{0.5\baselineskip}

\begin{center}
\small
\setlength{\tabcolsep}{3pt}
\captionof{table}{Sensor positions in the experimental channel (coordinates in meters). Sensor 16 was not installed.}
\label{tab:sensor_positions}
\begin{tabular}{@{}cccc@{\hspace{6pt}}cccc@{}}
\toprule
ID & $x$ & $y$ & $z$ & ID & $x$ & $y$ & $z$ \\
\midrule
1  & 0.77 & 0.24 & 0.80 & 17 & 0.46 & 4.39 & 0.80 \\
2  & 0.46 & 0.00 & 0.80 & 18 & 0.47 & 4.63 & 0.52 \\
3  & 0.48 & 0.25 & 0.52 & 19 & 0.14 & 4.68 & 0.52 \\
4  & 0.13 & 0.26 & 0.80 & 20 & 0.75 & 0.24 & 0.00 \\
5  & 0.48 & 1.33 & 0.52 & 21 & 0.45 & 0.00 & 0.00 \\
6  & 0.46 & 1.11 & 0.80 & 22 & 0.16 & 0.24 & 0.00 \\
7  & 0.90 & 2.22 & 0.80 & 23 & 0.47 & 0.24 & 0.28 \\
8  & 0.47 & 2.46 & 0.52 & 24 & 0.46 & 1.33 & 0.27 \\
9  & 0.46 & 2.22 & 0.80 & 25 & 0.47 & 1.10 & 0.00 \\
10 & 0.74 & 2.45 & 0.80 & 26 & 0.46 & 2.46 & 0.27 \\
11 & 0.15 & 2.44 & 0.80 & 27 & 0.47 & 2.23 & 0.00 \\
12 & 0.00 & 2.22 & 0.80 & 28 & 0.47 & 3.56 & 0.27 \\
13 & 0.47 & 3.54 & 0.52 & 29 & 0.47 & 4.63 & 0.27 \\
14 & 0.46 & 3.31 & 0.80 & 30 & 0.46 & 4.44 & 0.00 \\
15 & 0.78 & 4.65 & 0.80 &    &      &      &      \\
\bottomrule
\end{tabular}
\end{center}

\vspace{0.5\baselineskip}

\begin{center}
\small
\setlength{\tabcolsep}{3pt}
\captionof{table}{Dataset split and distribution for exposed sensors (H$_2$ $>0.009$). Training/validation: 80/20 stratified by mass flow. Test: held out.}
\label{tab:dataset_split}
\begin{tabular}{@{}lccc@{}}
\toprule
& Train. & Val. & Test \\
\midrule
Records & 78,133 & 23,479 & 14,694 \\
\midrule
\multicolumn{4}{@{}l}{\textit{Mass flow (kg/s)}} \\
\quad Min & 0.029 & 0.086 & 0.093 \\
\quad Max & 1.270 & 1.227 & 1.250 \\
\quad Mean & 0.401 & 0.540 & 0.381 \\
\quad Std & 0.283 & 0.365 & 0.468 \\
\midrule
\multicolumn{4}{@{}l}{\textit{H$_2$ concentration}} \\
\quad Min & 9.00e-3 & 9.00e-3 & 9.00e-3 \\
\quad Max & 0.244 & 0.308 & 0.277 \\
\quad Mean & 0.082 & 0.103 & 0.100 \\
\quad Std & 0.039 & 0.061 & 0.049 \\
\bottomrule
\end{tabular}
\end{center}

\vspace{0.5\baselineskip}

\begin{center}
\small
\setlength{\tabcolsep}{3pt}
\captionof{table}{Validation performance of VNNGP models with the full additive kernel \eqref{eq:general_additive} on the exposed-only dataset. All kernels are RBF except where noted. Metrics correspond to the best checkpoint.}
\label{tab:scaleAdditive_twoModes_results}
\begin{tabular}{@{}lcccccc@{}}
\toprule
Kernel & $k$ & LR & Likelihood & MAE & RMSE & Loss \\
\midrule
RBF & 32 & 0.008 & Beta & 0.0944 & 0.1075 & 0.1816 \\
RBF & 64 & 0.008 & Beta & 0.0944 & 0.1076 & 0.2791 \\
RBF & 128 & 0.006 & Beta & 0.4148 & 0.4161 & 1.2120 \\
\midrule
RBF & 32 & 0.008 & Gaussian & 0.0280 & 0.0385 & $-$0.1021 \\
RBF & 64 & 0.008 & Gaussian & 0.0274 & 0.0384 & 0.1662 \\
RBF & 64 & 0.005 & Gaussian & 0.0263 & 0.0312 & 0.8374 \\
RBF & 64 & 0.009 & Gaussian & 0.0201 & 0.0277 & $-$0.0166 \\
Matern & 64 & 0.008 & Gaussian & 0.0261 & 0.0315 & 1.3421 \\
RBF & 128 & 0.008 & Gaussian & 0.0167 & 0.0223 & 2.3128 \\
\bottomrule
\end{tabular}
\end{center}

\vspace{0.5\baselineskip}

\begin{center}
\small
\setlength{\tabcolsep}{4pt}
\captionof{table}{Validation performance of VNNGP models with the light additive kernel \eqref{eq:light_additive_kernel} on the exposed-only dataset. Metrics correspond to the best checkpoint.}
\label{tab:light_Additive_twoModes_results}
\begin{tabular}{@{}lcccccc@{}}
\toprule
Kernel & $k$ & LR & Likelihood & MAE & RMSE & Loss \\
\midrule
RBF & 16 & 0.008 & Beta & 0.4206 & 0.4222 & 0.8641 \\
\midrule
RBF & 16 & 0.009 & Gaussian & 0.0136 & 0.0193 & 1.3804 \\
RBF & 32 & 0.005 & Gaussian & 0.0145 & 0.0200 & 0.5485 \\
RBF & 32 & 0.008 & Gaussian & 0.0146 & 0.0201 & 0.8297 \\
RBF & 64 & 0.005 & Gaussian & 0.0155 & 0.0207 & 0.7672 \\
RBF & 128 & 0.005 & Gaussian & 0.0166 & 0.0220 & 0.8357 \\
RBF & 256 & 0.008 & Gaussian & 0.0156 & 0.0210 & 1.5271 \\
\bottomrule
\end{tabular}
\end{center}

\vspace{0.5\baselineskip}

\begin{center}
\small
\setlength{\tabcolsep}{3pt}
\captionof{table}{Validation performance of SVGP models with the light additive kernel \eqref{eq:light_additive_kernel} on the exposed-only dataset. $^{\dagger}$Learning rate $0.085$; all others $0.008$.}
\label{tab:light_Additive_SVGP_val}
\begin{tabular}{@{}llccccc@{}}
\toprule
Model & Kernel & $M$ & Likelihood & MAE & RMSE & Loss \\
\midrule
SVGP & RBF & 4000 & Beta & 0.0128 & 0.0169 & $-2.5602$ \\
SVGP$^{\dagger}$ & RBF & 6000 & Beta & 0.0179 & 0.0235 & $-2.6051$ \\
SVGP & Mat.\ 1/2 & 4000 & Beta & 0.0141 & 0.0179 & $-2.3376$ \\
SVGP & Mat.\ 3/2 & 4000 & Beta & 0.0172 & 0.0209 & $-2.2477$ \\
\midrule
SVGP & RBF & 4000 & Gaussian & 0.0119 & 0.0160 & $-0.1836$ \\
SVGP & RBF & 5000 & Gaussian & 0.0116 & 0.0156 & $-0.2309$ \\
SVGP & RBF & 6000 & Gaussian & 0.0120 & 0.0162 & $-0.2287$ \\
\bottomrule
\end{tabular}
\end{center}

\vspace{0.5\baselineskip}

\begin{center}
\includegraphics[width=0.9\linewidth]{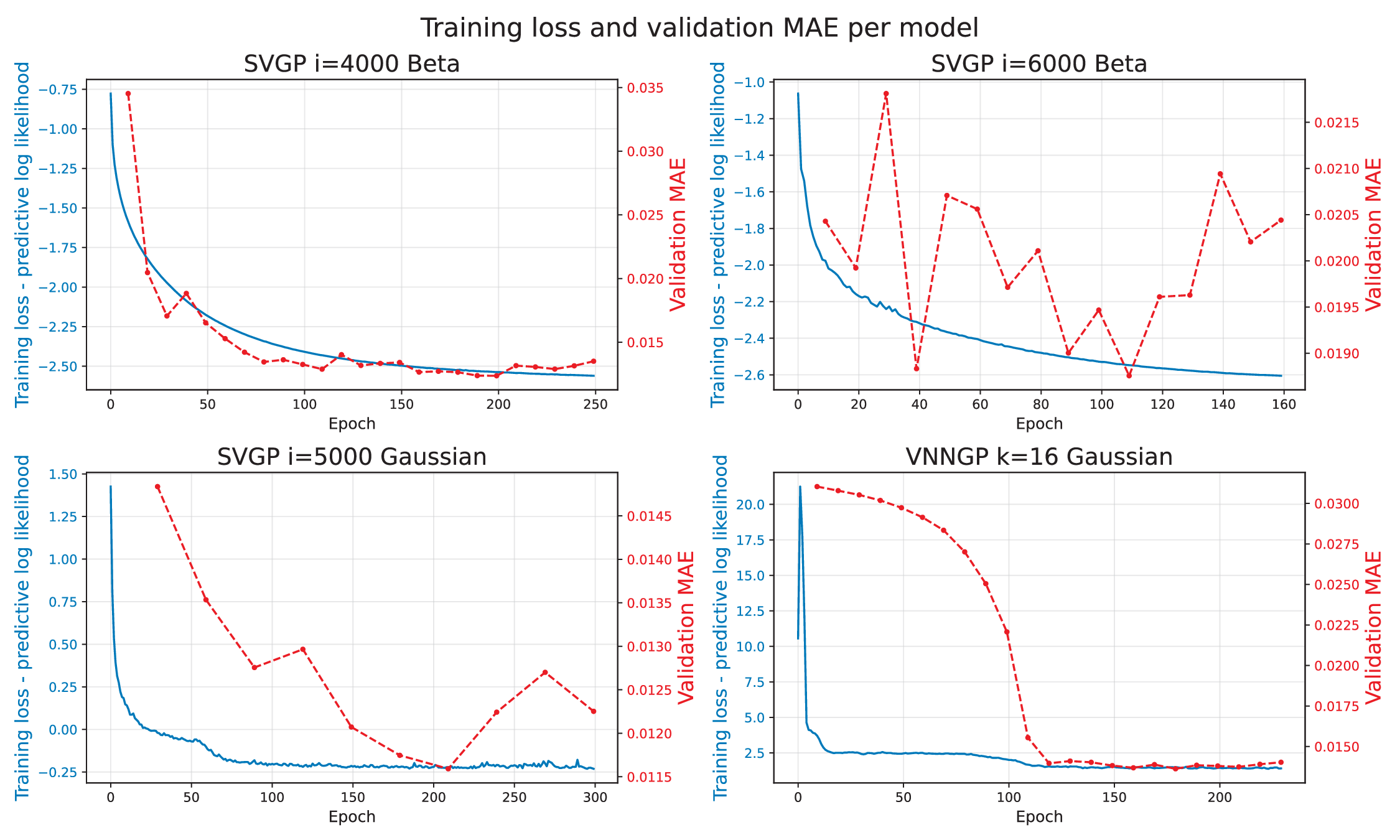}
\captionof{figure}{Training loss and MAE validation loss for four of the best models.}
\label{fig:loss_best_models}
\end{center}

\vspace{0.5\baselineskip}

\begin{center}
\small
\setlength{\tabcolsep}{4pt}
\captionof{table}{Test performance of the best models with the light additive kernel \eqref{eq:light_additive_kernel}. $^{*}$Learning rate $0.085$; all others $0.008$.}
\label{tab:light_Additive_test_perf}
\begin{tabular}{@{}lcccc@{}}
\toprule
Model & Likelihood & MAE & RMSE & NLL \\
\midrule
SVGP$_{4000}$ & Gaussian & 0.0210 & 0.0263 & 13.6988 \\
SVGP$_{5000}$ & Gaussian & 0.0211 & 0.0262 & 13.8401 \\
SVGP$_{6000}$ & Gaussian & 0.0207 & 0.0255 & 12.6712 \\
\midrule
VNNGP$_{K16}$ & Gaussian & 0.0248 & 0.0359 & 17.5418 \\
\midrule
SVGP$_{4000}$ & Beta & 0.0209 & 0.0267 & $-1.2255$ \\
SVGP$_{6000}^{*}$ & Beta & 0.0176 & 0.0226 & $-1.5340$ \\
\bottomrule
\end{tabular}
\end{center}

\vspace{0.5\baselineskip}

\begin{center}
\includegraphics[width=0.9\linewidth]{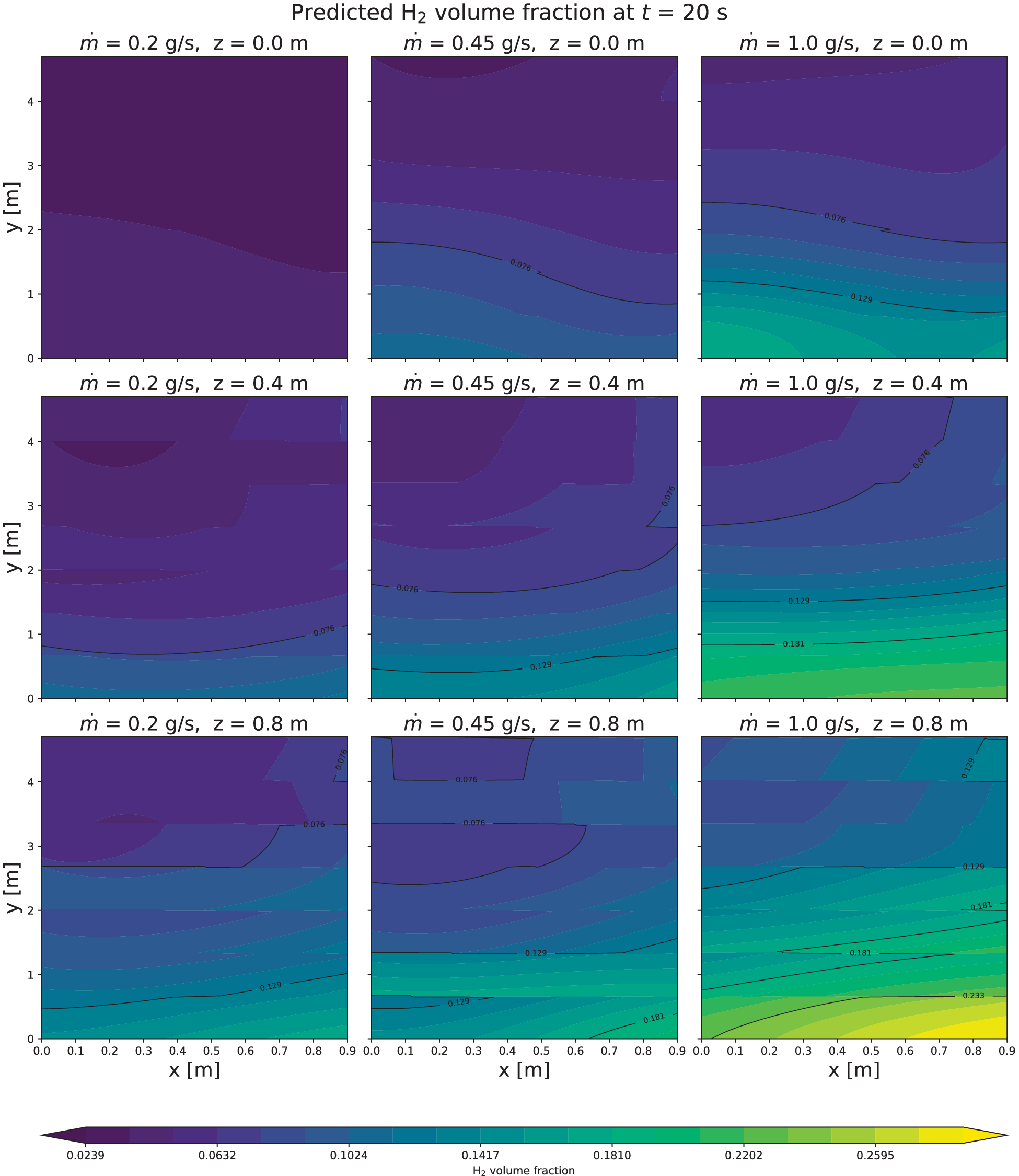}
\captionof{figure}{Predicted $H_2$ volume fraction at $t=20$\,s using the final SVGP model (Beta likelihood, $M=6000$) for leakage rates $\hat{m} \in \{0.2, 0.45, 1.0\}$\,g/s at heights $z \in \{0.0, 0.4, 0.8\}$\,m.}
\label{fig:contour}
\end{center}

\FloatBarrier
\printcredits

\bibliographystyle{cas-model2-names}
\FloatBarrier
\bibliography{cas-refs}

\end{document}